\documentclass{article}

\usepackage{times}
\usepackage{graphicx} 
\usepackage{subfigure} 

\usepackage{framed}

\usepackage{natbib}

\usepackage{algorithm}
\usepackage{algorithmic}

\usepackage{hyperref}

\usepackage{graphics, subfigure, theorem, times, amsfonts, amsmath, amssymb, cite, verbatim, comment}
\usepackage{tikz}
\usepackage{framed}
\usetikzlibrary{shapes,arrows}
\tikzstyle{phantom vertex} = [ ellipse, 
                               anchor = center, 
                               minimum height = 1*\unit, 
                               minimum width  = 1*\unit,
                               inner sep=0pt,
                               anchor=center]
\tikzstyle{red vertex}   = [black, fill = red!20,   phantom vertex, draw]
\tikzstyle{black vertex} = [black, fill = black!20, phantom vertex, draw]
\tikzstyle{blue vertex}  = [black, fill = blue!20,  phantom vertex, draw]
\tikzstyle{green vertex} = [black, fill = green!20,  phantom vertex, draw]
\tikzstyle{vertex}       = [draw, phantom vertex]

\tikzstyle{point} = [ellipse, inner sep=0pt, draw, fill=white, anchor = center,
                     minimum height = 0.05*\unit, minimum width  = 0.05*\unit]

\usepackage{pgfplots}
\usepackage{color}
\usepackage{needspace}

\newcommand{\myindentedparagraph}[1]{\needspace{1\baselineskip}\smallskip \hangindent=11pt \hangafter=0 \noindent{\it #1.}}

\newenvironment{indentedparagraph}[1] 
{\begin{list}{}%
         {\setlength{\leftmargin}{11pt}
          \setlength{\topsep}{10pt}}
         \item[]{\it #1.}}
{\end{list}}

\usepackage{flushend}
\usepackage{multirow}
\usepackage{rotating}
\usepackage{appendix}
\input{mysymbol.sty}

\newtheorem{lemma}{\hspace{0pt}\bf Lemma}

\newtheorem{proposition}{\hspace{0pt}\bf Proposition}

\newtheorem{theorem}{\hspace{0pt}\bf Theorem}

\newtheorem{remark}{\hspace{0pt}\bf Remark}

\newtheorem{definition}{\hspace{0pt}\bf Definition}

\def \sep {\text{\normalfont sep}}

\newcommand{\dN}[2]{d_{\mathcal N}({#1},{#2})}
\newcommand{\QED}{\hfill\ensuremath{\blacksquare}}

\usepackage[accepted]{icml2014}

\icmltitlerunning{Hierarchical Quasi-Clustering Methods for Asymmetric Networks}

\begin{document} 

\twocolumn[
\icmltitle{Hierarchical Quasi-Clustering Methods for Asymmetric Networks}

\icmlauthor{Gunnar Carlsson}{gunnar@math.stanford.edu}
\icmladdress{Department of Mathematics, Stanford University}
\icmlauthor{Facundo M\'emoli}{memoli@math.osu.edu}
\icmladdress{Department of Mathematics and Department of Computer Science and Engineering, Ohio State University}
\icmlauthor{Alejandro Ribeiro}{aribeiro@seas.upenn.edu}
\icmlauthor{Santiago Segarra}{ssegarra@seas.upenn.edu}
\icmladdress{Department of Electrical and Systems Engineering, University of Pennsylvania}

\icmlkeywords{clustering, asymmetric networks}

\vskip 0.3in
]

\begin{abstract} 
This paper introduces hierarchical quasi-clustering methods, a generalization of hierarchical clustering for asymmetric networks where the output structure preserves the asymmetry of the input data. We show that this output structure is equivalent to a finite quasi-ultrametric space and study admissibility with respect to two desirable properties. We prove that a modified version of single linkage is the only admissible quasi-clustering method. Moreover, we show stability of the proposed method and we establish invariance properties fulfilled by it. Algorithms are further developed and the value of quasi-clustering analysis is illustrated with a study of internal migration within United States.
\end{abstract} 


\section{Introduction}
\label{sec_introduction}

Given a network of interactions, hierarchical clustering methods determine a dendrogram, i.e. a family of nested partitions indexed by a resolution parameter. Clusters that arise at a given resolution correspond to sets of nodes that are more similar to each other than to the rest and, as such, can be used to study the formation of groups and communities \cite{ShiMalik00, GirvanNewman02, GirvanNewman04, spectral-clustering, NgEtal02, lance67general, clusteringref}. For asymmetric networks, in which the dissimilarity from node $x$ to node $x'$ may differ from the one from $x'$ to $x$ \cite{SaitoYadohisa04}, the determination of said clusters is not a straightforward generalization of the methods used to cluster symmetric datasets \cite{hubert-min,slater1976hierarchical,boyd-asymmetric,tarjan-improved,slater1984partial,murtagh-multidimensional,PentneyMeila05, MeilaPentney07, ZhaoKarypis05}. 

This difficulty motivates formal developments whereby hierarchical clustering methods are constructed as those that are admissible with respect to some reasonable properties \cite{clust-um, CarlssonMemoli10, Carlssonetal13}. A fundamental distinction between symmetric and asymmetric networks is that while it is easy to obtain uniqueness results for the former \cite{clust-um}, there are a variety of methods that are admissible for the latter \cite{Carlssonetal13}. Although one could conceive of imposing further restrictions to winnow the space of admissible methods for clustering asymmetric networks, it is actually reasonable that multiple methods should exist. Since dendrograms are symmetric structures one has to make a decision as to how to derive symmetry from an asymmetric dataset and there are different stages of the clustering process at which such symmetrization can be carried out \cite{Carlssonetal13}. In a sense, there is a fundamental mismatch between having a network of {\it asymmetric} relations as input and a {\it symmetric} dendrogram as output.

This paper develops a generalization of dendrograms and hierarchical clustering methods to allow for asymmetric output structures. We refer to these asymmetric structures as quasi-dendrograms and to the procedures that generate them as hierarchical quasi-clustering methods. Since the symmetry in dendrograms can be traced back to the symmetry of equivalence relations we start by defining a quasi-equivalence relation as one that is reflexive and transitive but not necessarily symmetric (Section \ref{sec_full_characterization_asymmetric}). We then define a quasi-partition as the structure induced by a quasi-equivalence relation, a quasi-dendrogram as a nested collection of quasi-partitions, and a hierarchical quasi-clustering method as a map from the space of networks to the space of quasi-dendrograms (Section \ref{sec_quasi_dendrograms}). Quasi-partitions are similar to regular partitions in that they contain disjoint blocks of nodes but they also include an influence structure between the blocks derived from the asymmetry in the original network. This influence structure defines a partial order over the blocks \cite{Harzheim05}. 

We proceed to study admissibility of quasi-clustering methods with respect to the directed axioms of value and transformation. The Directed Axiom of Value states that the quasi-clustering of a network of two nodes is the network itself. The Directed Axiom of Transformation states that reducing dissimilarities cannot lead to looser quasi-clusters. We show that there is a unique quasi-clustering method admissible with respect to these axioms and that this method is an asymmetric version of the single linkage clustering method (Section \ref{sec_existance_uniqueness_quasi_clustering}). The analysis in this section hinges upon an equivalence between quasi-dendrograms and quasi-ultrametrics (Section \ref{sec_quasi_ultrametrics}) that generalizes the well-known equivalence between dendrograms and ultrametrics \cite{jardine-sibson}. 

Exploiting the fact that quasi-dendrograms can be represented by quasi-ultrametrics, we propose a quantitative notion of stability of quasi-clustering methods (Section \ref{sec:stab}). We prove that the unique method from Section \ref{sec_existance_uniqueness_quasi_clustering} is stable in the sense that we propose. We also establish several invariance properties enjoyed by this method.

In order to apply the quasi-clustering method to real data, we derive an algorithm based on matrix powers in a dioid algebra \cite{GondranMinoux08} (Section \ref{sec_algorithms}).
As an example, we cluster a network that contains information about the internal migration between states of the United States for the year 2011 (Section \ref{sec_applications}). The quasi-clustering output unveils that migration is dominated by geographical proximity. Moreover, by exploiting the asymmetric influence between clusters, one can show the migrational influence of California over the West Coast. 

Proofs of results in this paper not contained in the main body can be found in the supplementary material.

\section {Preliminaries}\label{sec_preliminaries}

A network $N$ is a pair $(X, A_X)$ where $X$ is a finite set of points or nodes and $A_X: X \times X \to \reals_+$ is a dissimilarity function. The value $A_X(x,x')$ is assumed to be non-negative for all pairs $(x,x') \in X \times X$ and 0 if and only if $x=x'$. However, $A_X$ need not satisfy the triangle inequality and, more consequential for the problem considered here, may be asymmetric in that it is possible to have $A_X(x,x')\neq A_X(x',x)$ for some $x \neq x'$. We further define $\ccalN$ as the set of all networks. Networks $N\in\ccalN$ can have different node sets $X$ and different dissimilarities $A_X$.

A conventional non-hierarchical clustering of the set $X$ is a partition $P$, i.e., a collection of sets $P=\{B_1,\ldots, B_J\}$ which are pairwise disjoint, $B_i\cap B_j =\emptyset$ for $i\neq j$, and required to cover $X$, $\cup_{i=1}^{J} B_i = X$. The sets $B_1, B_2, \ldots B_J$ are called the \emph{blocks} of $P$ and represent \emph{clusters}. A partition $P=\{B_1,\ldots, B_J\}$ of $X$ induces and is induced by an equivalence relation $\sim $ on $X$ such that for all $x, x', x'' \in X$ we have that $x \sim x$, $x\sim x'$ if and only if $x'\sim x$, and $x\sim x'$ combined with $x'\sim x''$ implies $x\sim x''$. In hierarchical clustering methods the output is not a single partition $P$ but a nested collection $D_X$ of partitions $D_X(\delta)$ of $X$ indexed by a resolution parameter $\delta\geq0$. For a given $D_X$, we say that two nodes $x$ and $x'$ are equivalent at resolution $\delta \geq 0$ and write $x\sim_{D_X(\delta)} x'$ if and only if nodes $x$ and $x'$ are in the same cluster of $D_X(\delta)$. The nested collection $D_X$ is termed a \emph{dendrogram} \cite{jardine-sibson}. The interpretation of a dendrogram is that of a structure which yields different clusterings at different resolutions. At resolution $\delta=0$ each point is in a cluster of its own and as the resolution parameter $\delta$ increases, nodes start forming clusters. We denote  by $[x]_{\delta}$ the equivalence class to which the node $x \in X$ belongs at resolution $\delta$, i.e. $[x]_\delta := \{x' \in X \given x \sim_{D_X(\delta)} x'\}$. 

In our development of hierarchical quasi-clustering methods, the concepts of \emph{chain} and \emph{chain cost} are important. Given a network $(X, A_X)$ and $x, x' \in X$, a chain $C(x, x')$ is an \emph{ordered} sequence of nodes in $X$, 
\begin{equation}\label{eqn_definition_chain}
   C(x, x')=[x=x_0, x_1, \ldots , x_{l-1}, x_l=x'],
\end{equation}
which starts at $x$ and ends at $x'$. We say that $C(x, x')$ links or connects $x$ to $x'$. 
The \emph{links} of a chain are the edges connecting consecutive nodes of the chain in the direction given by the chain. We define the \emph{cost} of a chain \eqref{eqn_definition_chain} as the maximum dissimilarity $\max_{i | x_i\in C(x,x')}A_X(x_i,x_{i+1})$ encountered when traversing its links in order. 


\section{Quasi-Clustering methods}\label{sec_full_characterization_asymmetric}

A partition $P=\{B_1,\ldots, B_J\}$ of a set $X$ can be interpreted as a reduction in data complexity in which variations between elements of a group are neglected in favor of the larger dissimilarities between elements of different groups. This is natural when clustering datasets endowed with symmetric dissimilarities because the concepts of a node $x\in X$ being close to another node $x'\in X$ and $x'$ being close to $x$ are equivalent. In an asymmetric network these concepts are different and this difference motivates the definition of structures more general than partitions. 

Considering that a partition $P=\{B_1,\ldots, B_J\}$ of $X$ is induced by an equivalence relation $\sim$ on $X$ we search for the equivalent of an asymmetric partition by removing the symmetry property in the definition of the equivalence relation. Thus, we define a \emph{quasi-equivalence} $\leadsto$ as a binary relation that satisfies the reflexivity and transitivity properties but is not necessarily symmetric as stated next. 

\begin{definition}\label{def_quasi_equivalence}
A binary relation $\leadsto$ between elements of a set $X$ is a quasi-equivalence if and only if the following properties hold true for all $x, x', x'' \in X$:
\begin{mylist}
\item[{\it (i) Reflexivity.}] Points are quasi-equivalent to themselves, $x \leadsto x$.
\item[{\it (ii) Transitivity.}] If $x\leadsto x'$ and $x'\leadsto x''$ then $x\leadsto x''$.
\end{mylist} \end{definition}

Quasi-equivalence relations are more often termed preorders or quasi-orders in the literature \cite{Harzheim05}. We choose the term quasi-equivalence to emphasize that they are a modified version of an equivalence relation.

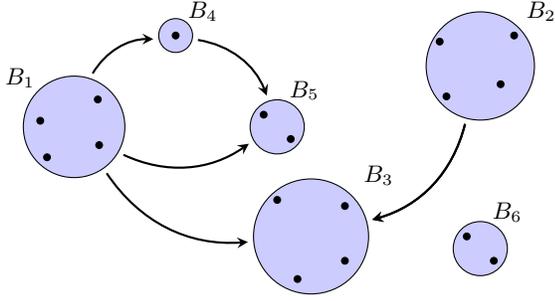
\begin{figure}
\centering
\def \thisplotscale {0.9}
\def \unit {\thisplotscale cm}

{\footnotesize
\begin{tikzpicture}[-stealth, shorten >=2,  shorten <=2, x = \unit, y=0.9*\unit]

    \node [blue vertex, minimum size = 1.5*\unit] at (0,0.1) (b1) {};
    \path (b1) ++ (-0.8,0.8) node {\small $B_1$};
    \node [point, minimum height = 0.1*\unit, minimum width  = 0.1*\unit, fill=black] at ( 0.37,-0.20) (1) {}; 
    \node [point, minimum height = 0.1*\unit, minimum width  = 0.1*\unit, fill=black] at ( 0.35, 0.55) (2) {};
    \node [point, minimum height = 0.1*\unit, minimum width  = 0.1*\unit, fill=black] at (-0.5, 0.2) (3) {}; 
    \node [point, minimum height = 0.1*\unit, minimum width  = 0.1*\unit, fill=black] at (-0.4,-0.4) (4) {}; 

    \path (b1) ++ (6.0,1.0) node [blue vertex, minimum size = 1.6*\unit] (b2) {}; 
    \path (b2) ++ (0.9,0.9) node {\small $B_2$};    
    \path (b2) ++ (-0.6,0.4) node [point, minimum height = 0.1*\unit, minimum width  = 0.1*\unit, fill=black] (1p) {};  \path (b2) ++ (0.3,-0.3) node [point, minimum height = 0.1*\unit, minimum width  = 0.1*\unit, fill=black] (2p) {};
    \path (b2) ++ (0.5,0.5) node [point, minimum height = 0.1*\unit, minimum width  = 0.1*\unit, fill=black] (3p) {};  \path (b2) ++ (-0.5,-0.5) node [point, minimum height = 0.1*\unit, minimum width  = 0.1*\unit, fill=black] (4p) {};
                            (4p) edge (1p) (2p) edge (1p) (4p) edge (3p) ;
     
    \path (b1) ++ (3.5,-1.8) node [blue vertex, minimum size = 1.7*\unit] (b3) {}; 
    \path (b3) ++ (1,1) node {\small $B_3$};    
    \path (b3) ++ (-0.5,0.6) node [point, minimum height = 0.1*\unit, minimum width  = 0.1*\unit, fill=black] (1pp) {};  \path (b3) ++ (0.5,-0.4) node [point, minimum height = 0.1*\unit, minimum width  = 0.1*\unit, fill=black] (2pp) {};
    \path (b3) ++ (0.5,0.5) node [point, minimum height = 0.1*\unit, minimum width  = 0.1*\unit, fill=black] (3pp) {};  \path (b3) ++ (-0.2,-0.7) node [point, minimum height = 0.1*\unit, minimum width  = 0.1*\unit, fill=black] (4pp) {};
   
    \path (b1) ++ (1.5,1.5) node [blue vertex, minimum size = 0.5*\unit] (b4) {}; 
    \path (b4) ++ (+0.4,0.4) node {\small $B_4$};    
    \path (b4) ++ (0,0) node [point, minimum height = 0.1*\unit, minimum width  = 0.1*\unit, fill=black] (1ppp) {}; 

    \path (b1) ++ (3,0) node [blue vertex, minimum size = 0.8*\unit] (b5) {}; 
    \path (b5) ++ (+0.4,+0.6) node {\small $B_5$};    
    \path (b5) ++ (-0.2,0.2) node [point, minimum height = 0.1*\unit, minimum width  = 0.1*\unit, fill=black] (1pppp) {};  
    \path (b5) ++ (0.2,-0.2) node [point, minimum height = 0.1*\unit, minimum width  = 0.1*\unit, fill=black] (2pppp) {};

    \path (b1) ++ (6,-2) node [blue vertex, minimum size = 0.8*\unit] (b6) {}; 
    \path (b6) ++ (+0.4,+0.6) node {\small $B_6$};    
    \path (b6) ++ (-0.2,0.2) node [point, minimum height = 0.1*\unit, minimum width  = 0.1*\unit, fill=black] (1pppp) {};  
    \path (b6) ++ (0.2,-0.2) node [point, minimum height = 0.1*\unit, minimum width  = 0.1*\unit, fill=black] (2pppp) {};

    \path (b1) edge [thick, bend right, above]  node {} (b3);
    \path (b1) edge [thick, bend left, above]  node {} (b4);    
    \path (b1) edge [thick, bend right, above] node {} (b5);    
    \path (b2) edge [thick, bend left, pos=0.52, right] node {} (b3);    
    \path (b2) edge [thick, bend left, pos=0.52, right] node {} (b3);            
    \path (b4) edge [thick, bend left, pos=0.52, right] node {} (b5);   
\end{tikzpicture}}
\caption{A quasi-partition $\tdP=(P,E)$ on a set of nodes. The vertex set $P$ of the quasi-partition is given by a partition of the nodes $P=\{B_1, B_2, \ldots, B_6\}$. The edges of the directed graph $\tdP=(P,E)$ represent unidirectional influence between the blocks of the partition.}
\label{fig_quasi_partition_example}
\end{figure}

We define a \emph{quasi-partition} of the set $X$ as a directed, unweighted graph $\tdP=(P,E)$ with no self-loops where the vertex set $P$ is a partition $P=\{B_1,\ldots, B_J\}$ of the space $X$ and the edge set $E \subseteq P \times P$ is such that the following properties are satisfied (see Fig. \ref{fig_quasi_partition_example}):

\myindentedparagraph{(QP1) Unidirectionality} For any given pair of distinct blocks $B_i$, $B_j \in P$ we have at most one edge between them. Thus, if for some $i \neq j$ we have $(B_i,B_j)\in E$ then $(B_j,B_i)\notin E$.

\myindentedparagraph{(QP2) Transitivity} If there are edges between blocks $B_i$ and $B_j$ and between blocks $B_j$ and $B_k$, then there is an edge between blocks $B_i$ and $B_k$. 

\medskip\noindent The vertex set $P$ of a quasi-partition $\tdP=(P,E)$ represents sets of nodes that can influence each other, whereas the edges in $E$ capture the notion of directed influence from one group to the next. In the example in Fig. \ref{fig_quasi_partition_example}, nodes which are drawn together can exert influence on each other. This gives rise to the blocks $B_i$ which form the vertex set $P$ of the quasi-partition. Additionally, some blocks have influence over others in only one direction. E.g., block $B_1$ can influence $B_4$ but not vice versa. This latter fact motivates keeping $B_1$ and $B_4$ as separate blocks in the partition whereas the former motivates the addition of the directed influence edge $(B_1,B_4)$. Likewise, $B_1$ can influence $B_3$, $B_2$ can influence $B_3$ and $B_4$ can influence $B_5$ but none of these influences are true in the opposite direction. Block $B_1$ need not be able to directly influence $B_5$, but can influence it through $B_4$, hence the edge from $B_1$ to $B_5$, in accordance with (QP2). All other influence relations are not meaningful, justifying the lack of connections between the other blocks. Observe that there are no bidirectional edges as required by (QP1).

Requirements (QP1) and (QP2) in the definition of quasi-partition represent the relational structure that emerges from quasi-equivalence relations as we state in the following proposition.

\begin{proposition}\label{prop_quasi_equiv_quasi_part}
Given a node set $X$ and a quasi-equivalence relation $\leadsto$ on $X$ [cf. Definition \ref{def_quasi_equivalence}] define the relation $\leftrightarrow$ on $X$ as
\begin{equation}\label{eqn_quasi_equiv_equiv}
x \leftrightarrow x' \quad \iff \quad x \leadsto x' \,\,\, \text{\normalfont and} \,\,\, x' \leadsto x,
\end{equation}
for all $x, x' \in X$. Then, $\leftrightarrow$ is an equivalence relation. Let $P = \{B_1, \ldots , B_J\}$ be the partition of $X$ induced by $\leftrightarrow$. Define $E \subseteq P \times P$ such that for all distinct $B_i, B_j \in P$
\begin{equation}\label{eqn_quasi_equiv_edges_quasi_partition}
(B_i, B_j) \in E \quad \iff \quad x_i \leadsto x_j,
\end{equation}
for some $x_i \in B_i$ and $x_j \in B_j$. Then, $\tdP=(P,E)$ is a quasi-partition of $X$. Conversely, given a quasi-partition $\tdP=(P,E)$ of $X$, define the binary relation $\leadsto$ on $X$ so that for all $x, x' \in X$
\begin{equation}
x \leadsto x' \iff [x] = [x'] \,\,\, \text{or} \,\,\, ([x], [x']) \in E,
\end{equation}
where $[x] \in P$ is the block of the partition $P$ that contains the node $x$ and similarly for $[x']$. Then, $\leadsto$ is a quasi-equivalence on $X$.
\end{proposition}
\begin{myproof}
See Theorem 4.9, Ch. 1.4 in \cite{Harzheim05}.
\end{myproof}

In the same way that an equivalence relation induces and is induced by a partition on a given node set $X$, Proposition \ref{prop_quasi_equiv_quasi_part} shows that a quasi-equivalence relation induces and is induced by a quasi-partition on $X$. We can then adopt the construction of quasi-partitions as the natural generalization of clustering problems when given asymmetric data. Further, observe that if the edge set $E$ contains no edges, $\tdP=(P,E)$ is equivalent to the regular partition $P$ when ignoring the empty edge set. In this sense, partitions are particular cases of quasi-partitions having the generic form $\tdP=(P,\emptyset)$. To allow generalizations of hierarchical clustering methods with asymmetric outputs we introduce the notion of \emph{quasi-dendrogram} in the following section.


\subsection{Quasi-dendrograms}\label{sec_quasi_dendrograms}

Given that a dendrogram is defined as a nested set of partitions, we define a \emph{quasi-dendrogram} $\tilde{D}_X$ of the set $X$ as a collection of nested quasi-partitions $\tilde{D}_X(\delta)=(D_X(\delta), E_X(\delta))$ indexed by a resolution parameter $\delta \geq 0$. Recall the definition of $[x]_\delta$ from Section \ref{sec_preliminaries}. Formally, for $\tilde{D}_X$ to be a quasi-dendrogram we require the following conditions:

\begin{indentedparagraph}{(\~D1) Boundary conditions} At resolution $\delta=0$ all nodes are in separate clusters with no edges between them and for some $\delta_0$ sufficiently large all elements of $X$ are in a single cluster,
\begin{align}\label{eqn_dendrogram_boundary_conditions}
   & \tdD_X(0)  = \Big ( \big\{ \{x\}, \, x\in X\big\},\ \emptyset \Big), \nonumber\\
   & \tdD_X(\delta_0) = \Big ( \{ X\}, \emptyset \Big) \quad \forsome\ \delta_0 \geq 0.
\end{align}\end{indentedparagraph}

\begin{indentedparagraph}{(\~D2) Equivalence hierarchy} For any pair of points $x,x'$ for which $x\sim_{D_X(\delta_1)} x'$ at resolution $\delta_1$ we must have $x\sim_{D_X(\delta_2)} x'$ for all resolutions $\delta_2 > \delta_1$. \end{indentedparagraph} 

\begin{indentedparagraph}{(\~D3) Influence hierarchy} If there is an edge $([x]_{\delta_1}, [x']_{\delta_1}) \in E_X(\delta_1)$ between the equivalence classes $[x]_{\delta_1}$ and $[x']_{\delta_1}$ of nodes $x$ and $x'$ at resolution $\delta_1$, at any resolution $\delta_2>\delta_1$ we either have $([x]_{\delta_2}, [x']_{\delta_2}) \in E_X(\delta_2)$ or $[x]_{\delta_2}=[x']_{\delta_2}$. \end{indentedparagraph}

\begin{indentedparagraph}{(\~D4) Right continuity} For all $\delta \geq 0$ there exists $\epsilon > 0$ such that $\tdD_X(\delta)=\tdD_X(\delta')$ for all $\delta'\in [\delta, \delta+\epsilon]$. \end{indentedparagraph}

\noindent Requirement (\~D1) states that for resolution $\delta=0$ there should be no influence between any pair of nodes and that, for a large enough resolution $\delta=\delta_0$, there should be enough influence between the nodes for all of them to belong to the same cluster. 
According to (\~D2), nodes become ever more clustered since once they join together in a cluster, they stay together in the same cluster for all larger resolutions. 
Condition (\~D3) states for the edge set the analogous requirement that (\~D2) states for the node set. If there is an edge present at a given resolution $\delta_1$, that edge should persist at coarser resolutions $\delta_2>\delta_1$ except if the groups linked by the edge merge in a single cluster. Requirement (\~D4) is a technical condition that ensures the correct definition of a hierarchical structure [cf. \eqref{eqn_theo_dendrograms_as_quasi_ultrametrics_10} below].

Comparison of (\~D1), (\~D2), and (\~D4) with the three properties defining a dendrogram \cite{clust-um} implies that given a quasi-dendrogram $\tilde{D}_X=(D_X, E_X)$ on a node set $X$, the component $D_X$ is a dendrogram on $X$. I.e, the vertex sets $D_X(\delta)$ of the quasi-partitions $(D_X(\delta), E_X(\delta))$ for varying $\delta$ form a nested set of partitions. Hence, if the edge set $E_X(\delta) = \emptyset$ for every resolution parameter, $\tilde{D}_X$ recovers the structure of the dendrogram $D_X$. Thus, quasi-dendrograms are a generalization of dendrograms, or, equivalently, dendrograms are particular cases of quasi-dendrograms with empty edge sets. Regarding dendrograms $D_X$ as quasi-dendrograms $(D_X,\emptyset)$ with empty edge sets, we have that the set of all dendrograms $\ccalD$ is a subset of $\tilde{\ccalD}$, the set of all quasi-dendrograms.

A hierarchical clustering method $\ccalH: \ccalN \to \ccalD$ is defined as a map from the space of networks $\ccalN$ to the space of dendrograms $\ccalD$. This motivates the definition of a hierarchical \emph{quasi-clustering} method as follows.
\begin{definition}
A hierarchical quasi-clustering method $\tilde{\ccalH}$ is defined as a map from the space of networks $\ccalN$ to the space of quasi-dendrograms $\tilde{\ccalD}$,
\begin{equation}\label{eqn:def_hierarchical_quasi_clustering_methods}
\tilde{\ccalH}: \ccalN \to \tilde{\ccalD}.
\end{equation}
\end{definition}
Since $\ccalD \subset \tilde{\ccalD}$ we have that every clustering method is a quasi-clustering method but not vice versa. Our goal here is to study quasi-clustering methods satisfying desirable axioms that define the concept of admissibility. In order to facilitate this analysis, we introduce quasi-ultrametrics as asymmetric versions of ultrametrics and show their equivalence to quasi-dendrograms in the following section.

\begin{remark} \normalfont
Unidirectionality (QP1) ensures that no cycles containing exactly two nodes can exist in any quasi-partition $\tdP=(P,E)$. If there were longer cycles, transitivity (QP2) would imply that every two distinct nodes in a longer cycle would have to form a two-node cycle, contradicting (QP1). Thus, conditions (QP1) and (QP2) imply that every quasi-partition $\tdP=(P,E)$ is a directed acyclic graph (DAG). The fact that a DAG represents a partial order shows that our construction of a quasi-partition from a quasi-equivalence relation is consistent with the known set theoretic construction of a partial order on a partition of a set given a preorder on the set \cite{Harzheim05}. 
\end{remark}


\subsection{Quasi-ultrametrics}\label{sec_quasi_ultrametrics}
Given a node set $X$, a \emph{quasi-ultrametric} $\tdu_X$ on $X$ is a function $\tdu_X: X \times X \to \reals_+$ satisfying the identity property and the strong triangle inequality as we formally define next. 

\begin{definition}\label{def_quasi_ultrametric}
 Given a node set $X$, a quasi-ultrametric $\tdu_X$ is a non-negative function $\tdu_X: X \times X \to \reals_+$ satisfying the following properties for all $x,x',x''\in X.$:
\begin{mylist}
\item[{\it (i) Identity.}] $\tdu_X(x, x')=0$ if and only if $x=x'$.
\item[{\it (ii) Strong triangle inequality.}] $\tdu_X$ satisfies
\begin{equation}\label{eqn_def_strong_triangle_inequality}
\tdu_X(x, x') \leq \max ( \tdu_X(x, x''), \tdu_X(x'', x') ).
\end{equation}
\end{mylist} \end{definition}

\vspace{-0.2in}
Quasi-ultrametrics may be regarded as ultrametrics where the symmetry property is not imposed. In particular, the space $\tilde{\ccalU}$ of quasi-ultrametric networks, i.e. networks with quasi-ultrametrics as dissimilarity functions, is a superset of the space of ultrametric networks $\ccalU\subset\tilde{\ccalU}$. See \cite{gurvich} for a study of some structural properties of quasi-ultrametrics.

The following constructions and theorem establish a structure preserving equivalence between quasi-dendrograms and quasi-ultrametrics.

Consider the map $\Psi:\tilde{\mathcal{D}}\rightarrow\tilde{\mathcal{U}}$ defined as follows:
for a given quasi-dendrogram $\tilde{D}_X=(D_X, E_X)$ over the set $X$ write $\Psi(\tilde{D}_X) = (X,\tdu_X)$, where we define $\tdu_X(x,x')$ for each $x, x' \in X$ as the smallest resolution $\delta$ at which either both nodes belong to the same equivalence class $[x]_\delta=[x']_\delta$, i.e. $x \sim_{D_X(\delta)} x'$, or there exists an edge in $E_X(\delta)$ from the equivalence class $[x]_\delta$ to the equivalence class $[x']_\delta$,
\begin{align}\label{eqn_theo_dendrograms_as_quasi_ultrametrics_10}
   \tdu_X(x,x')& := \min \Big\{ \delta\geq 0 \, \Big| \\
   &  [x]_\delta = [x']_\delta \quad \text{\normalfont or} \quad ([x]_\delta, [x']_\delta) \in E_X(\delta) \Big\}. \nonumber
\end{align}

We also consider the map $\Upsilon:\tilde{\mathcal{U}}\rightarrow\tilde{\mathcal{D}}$ constructed as follows: for a given quasi-ultrametric $\tdu_X$ on the set $X$ and each $\delta \geq 0$ define the relation $\sim_{\tdu_X(\delta)}$ on $X$ as
\begin{equation}\label{eqn_theo_dendrograms_as_quasi_ultrametrics_20}
   x \sim_{\tdu_X(\delta)} x' \iff \max \big( \tdu_X(x,x'), \tdu_X(x',x) \big) \leq \delta.
\end{equation}
Define further $D_X(\delta) :=\big\{X \mod \sim_{\tdu_X(\delta)}\big\}$ and the edge set $E_X(\delta)$ for every $\delta \geq 0$ as follows: $B_1 \neq B_2 \in D_X(\delta)$ are such that 
\begin{equation}\label{eqn_theo_dendrograms_as_quasi_ultrametrics_30}
   (B_1, B_2) \in E_X(\delta) \iff \min_{\substack{x_1 \in B_1\\x_2 \in B_2}} \tdu_X(x_1, x_2) \leq \delta.
\end{equation}
\mbox{Finally, $\Upsilon(X, \tdu_X):= \tilde{D}_X$, where $\tilde{D}_X:=(D_X, E_X)$.}

\begin{theorem}\label{theo_equivalence_quasi_dendrogram_quasi_ultrametric}
The maps $\Psi:\tilde{\mathcal{D}}\rightarrow\tilde{\mathcal{U}}$ and $\Upsilon:\tilde{\mathcal{U}}\rightarrow\tilde{\mathcal{D}}$ are both well defined. Furthermore, $\Psi\circ\Upsilon$ is the identity on $\tilde{\mathcal{U}}$ and $\Upsilon\circ\Psi$ is the identity on $\tilde{\mathcal{D}}$.
\end{theorem}

Theorem \ref{theo_equivalence_quasi_dendrogram_quasi_ultrametric} implies that every quasi-dendrogram $\tilde{D}_X$ has an equivalent representation as a quasi-ultrametric network defined on the same underlying node set $X$.
This result allows us to reinterpret hierarchical quasi-clustering methods [cf. \eqref{eqn:def_hierarchical_quasi_clustering_methods}] as maps 
\begin{equation}
\tilde{\ccalH}:\ccalN \to \tilde{\ccalU},
\end{equation}
from the space of networks to the space of quasi-ultrametric networks. Apart from the 
theoretical importance of Theorem \ref{theo_equivalence_quasi_dendrogram_quasi_ultrametric}, this equivalence result is of practical importance since quasi-ultrametrics are mathematically more convenient to handle than quasi-dendrograms. Indeed, the results in this paper are derived in terms of quasi-ultrametrics. However, quasi-dendrograms are more convenient for representing data as illustrated in Section \ref{sec_applications}.

Given a quasi-dendrogram $\tdD_X=(D_X, E_X)$, the value $\tdu_X(x, x')$ of the associated quasi-ultrametric for $x, x' \in X$ is given by the minimum resolution $\delta$ at which $x$ can influence $x'$. This may occur when $x$ and $x'$ belong to the same block of $D_X(\delta)$ or when they belong to different blocks $B, B' \in D_X(\delta)$, but there is an edge from the block containing $x$ to the block containing $x'$, i.e. $(B, B') \in E_X(\delta)$. Conversely, given a quasi-ultrametric network $(X, \tdu_X)$, for a given resolution $\delta$ the graph $\tdD_X(\delta)$ has as a vertex set the classes of nodes whose quasi-ultrametric is less than $\delta$ in both directions. Furthermore, $\tdD_X(\delta)$ contains a directed edge between two distinct equivalence classes if the quasi-ultrametric from some node in the first class to some node in the second is not greater than $\delta$. 

\begin{figure}
\centering
\def \thisplotscale {0.44}
\def \unit {\thisplotscale cm}

{\small
\begin{tikzpicture}[-stealth, shorten >=2, scale = \thisplotscale]

    \node [blue vertex] at (1,3) (1) {$x_2$};
    \node [blue vertex] at (4,-1) (2) {$x_3$};    
    \node [blue vertex] at (-2,-1) (3) {$x_1$};

    \path (1) edge [bend left=20, above right] node {$3$} (2);	
    \path (2) edge [bend left=20, below] node {$2$} (3);
    \path (3) edge [bend left=20, above left] node {$1$} (1);    	

    \path (2) edge [bend left=20,left, pos=0.6] node {$1$} (1);	
    \path (3) edge [bend left=20, above] node {$3$} (2);
    \path (1) edge [bend left=20, right, , pos=0.4]  node {$2$} (3);    	
    

    \node [below] at (6,-3) {$\tilde{D}_X(\delta)$};

       \node [blue vertex, scale=0.75] at (-0.4,-5.2) (4) {$x_2$};
    \node [blue vertex, scale=0.75] at (1.1,-7) (5) {$x_3$};    
    \node [blue vertex, scale=0.75] at (-1.9,-7) (6) {$x_1$};
    
    
    \node [below] at (-0.4,-7.4) {$0 \leq \delta < 1$};

           \node [blue vertex, scale=0.75] at (4.4,-5.2) (4p) {$x_2$};
    \node [blue vertex, scale=0.75] at (5.9,-7) (5p) {$x_3$};    
    \node [blue vertex, scale=0.75] at (2.9,-7) (6p) {$x_1$};
    

       \path (6p) edge [bend left=20, above] node {} (4p);
    \path (5p) edge [bend right=20, right, , pos=0.4]  node {} (4p);    	
    
    \node [below] at (4.4,-7.4) {$1 \leq \delta < 2$};

     \node [blue vertex, scale=0.75] at (9.2,-5.2) (7p) {$x_{\{1,2\}}$};
      \node [blue vertex, scale=0.75] at (9.2,-7) (8p) {$x_3$};
      
      
         \path (8p) edge node {} (7p);

      \node [below] at (9.2,-7.4) {$2 \leq \delta < 3$};

         \node [blue vertex, scale=0.75] at (13,-6.1) (9p) {$x_{\{1,2, 3\}}$};
         
           \node [below] at (13,-7.4) {$\delta \geq 3$};




   \small
   \draw [-stealth] (6.5,-2) -- (6.5,3.5);
    \draw [-stealth] (6.3,-1.8) -- (13.3,-1.8);
    \draw [-, draw=black!30] (8,-1.8) -- (8,3.5);
      \draw [-, draw=black!30] (9.5,-1.8) -- (9.5,3.5);
       \draw [-, draw=black!30] (11,-1.8) -- (11,3.5);
    
   \draw[thick, -] (6.5, -1) -- ++(3,0) -- ++(0,1.5) -- ++(-3,0) ++(0,1.5) -- ++(4.5,0) -- ++(0,-2.25) -- ++(-1.5,0)-- ++(1.5,0)-- ++(0,1.125)-- ++(2,0);

    \node [below] at (13.3,-2.3) {$\delta$};
    \node [below] at (8,-2) {$1$};
    \node [below] at (9.5,-2) {$2$};
    \node [below] at (11,-2) {$3$};
    \node [left] at (6.5,-1) {$x_1$};
    \node [left] at (6.5,0.5) {$x_2$};
    \node [left] at (6.5,2) {$x_3$};
    
    \node [below] at (12,3.5) {$D_X$};
    
    \node [below] at (-2,4) {$\tdu_X$};
    
      \draw [-stealth] (-3,-8.5) -- (14,-8.5);
       \draw [-, draw=black!30] (-2.8,-8.5) -- (-2.8,-4.5);
       \node [below] at (-2.8, -8.5) {0};
        \draw [-, draw=black!30] (2,-8.5) -- (2,-4.5);
        \node [below] at (2,-8.5) {1};
         \draw [-, draw=black!30] (6.8,-8.5) -- (6.8,-4.5);
         \node [below] at (6.8,-8.5) {2};
          \draw [-, draw=black!30] (11.6,-8.5) -- (11.6,-4.5);
    \node [below] at (11.6,-8.5) {3};
       \node [below] at (13.6,-8.8) {$\delta$};
    
    
    

\end{tikzpicture}
}
\caption{Equivalence between quasi-dendrograms and quasi-ultrametrics. A quasi-ultrametric $\tdu_X$ is defined on three nodes and the equivalent quasi-dendrogram $\tilde{D}_X=(D_X, E_X)$.}
\vspace{-0.2in}
\label{fig_quasi_dendrogram_example}
\end{figure}
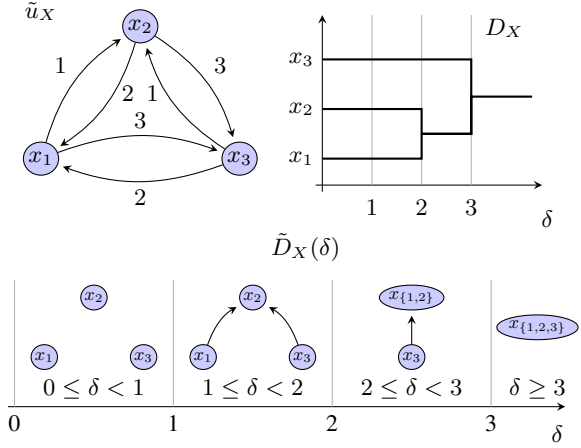

In Fig. \ref{fig_quasi_dendrogram_example} we present an example of the equivalence between quasi-dendrograms and quasi-ultrametric networks stated by Theorem \ref{theo_equivalence_quasi_dendrogram_quasi_ultrametric}. At the top left of the figure, we present a quasi-ultrametric $\tdu_X$ defined on a three-node set $X=\{x_1, x_2, x_3\}$. At the top right, we depict the dendrogram component $D_X$ of the quasi-dendrogram $\tilde{D}_X=(D_X, E_X)$ equivalent to $(X, \tdu_X)$ as given by Theorem \ref{theo_equivalence_quasi_dendrogram_quasi_ultrametric}. At the bottom of the figure, we present graphs $\tilde{D}_X(\delta)$ for a  range of resolutions $\delta \geq 0$.

To obtain $\tilde{D}_X$ from $\tdu_X$, we first obtain the dendrogram component $D_X$ by symmetrizing $\tdu_X$ to the maximum [cf. \eqref{eqn_theo_dendrograms_as_quasi_ultrametrics_20}], nodes $x_1$ and $x_2$ merge at resolution 2 and $x_3$ merges with $\{x_1, x_2\}$ at resolution 3. To see how the edges in $\tdD_X$ are obtained, at resolutions $0 \leq \delta < 1$, there are no edges since there is no quasi-ultrametric value between distinct nodes in this range [cf. \eqref{eqn_theo_dendrograms_as_quasi_ultrametrics_30}]. At resolution $\delta=1$, we reach the first non-zero values of $\tdu_X$ and hence the corresponding edges appear in $\tdD_X(1)$. At resolution $\delta=2$, nodes $x_1$ and $x_2$ merge and become the same vertex in graph $\tdD_X(2)$. Finally, at resolution $\delta=3$ all the nodes belong to the same equivalence class and hence $\tdD_X(3)$ contains only one vertex. Conversely, to obtain $\tdu_X$ from $\tilde{D}_X$ as depicted in the figure, note that at resolution $\delta=1$ two edges $([x_1]_1, [x_2]_1)$ and $([x_3]_1, [x_2]_1)$ appear in $\tdD_X(1)$, thus the corresponding values of the quasi-ultrametric are fixed to be $\tdu_X(x_1, x_2)=\tdu(x_3, x_2)=1$. At resolution $\delta=2$, when $x_1$ and $x_2$ merge into the same vertex in $\tdD_X(2)$, an edge is generated from $[x_3]_2$ to $[x_1]_2$ the equivalence class of $x_1$ at resolution $\delta=2$ which did not exist before, implying that $\tdu_X(x_3, x_1)=2$. Moreover, we have that $[x_2]_2 = [x_1]_2$, hence $\tdu_X(x_2, x_1)=2$. Finally, at $\tdD_X(3)$ there is only one equivalence class, thus the values of $\tdu_X$ that have not been defined so far must equal 3.

\subsection{Admissible quasi-clustering methods}\label{sec_admissibility}

We encode desirable properties of quasi-clustering methods into axioms which we use as a criterion for admissibility. The Directed Axiom of Value (\~A1) and the Directed Axiom of Transformation (\~A2) winnow the space of quasi-clustering methods by imposing conditions on their output quasi-ultrametrics which, by Theorem \ref{theo_equivalence_quasi_dendrogram_quasi_ultrametric}, is equivalent to imposing conditions on the output quasi-dendrograms. Defining an arbitrary two-node network $\vec{\Delta}_2(\alpha, \beta):= (\{p,q\}, A_{p,q})$ with $A_{p,q}(p,q)=\alpha$ and $A_{p,q}(q,p)=\beta$ for some $\alpha, \beta >0$,

\begin{indentedparagraph}{(\~A1) Directed Axiom of Value} $\tilde{\ccalH}(\vec{\Delta}_2(\alpha, \beta))= \vec{\Delta}_2(\alpha, \beta)$ for every two-node network $\vec{\Delta}_2(\alpha, \beta)$.
\end{indentedparagraph}

\begin{indentedparagraph}{(\~A2) Directed Axiom of Transformation} Consider two networks $N_X=(X,A_X)$ and $N_Y=(Y,A_Y)$ and a dissimilarity-reducing map $\phi:X\to Y$, i.e. a map $\phi$ such that for all $x,x' \in X$ it holds $A_X(x,x')\geq A_Y(\phi(x),\phi(x'))$. Then, for all $x, x' \in X$, the outputs $(X,\tdu_X)=\tilde{\ccalH}(X,A_X)$ and $(Y,\tdu_Y)=\tilde{\ccalH}(Y,A_Y)$ satisfy 
\begin{equation}\label{eqn_dissimilarity_reducing_quasi_ultrametric}
    \tdu_X(x,x') \geq \tdu_Y(\phi(x),\phi(x')).
\end{equation} \end{indentedparagraph}

The Directed Axiom of Transformation (\~A2) states that no influence relation can be weakened by a dissimilarity reducing transformation. That is, if relations in the network are strengthened, the tendency of nodes to cluster cannot decrease. The Directed Axiom of Value (\~A1) simply recognizes that in any two-node network, the dissimilarity function is itself a quasi-ultrametric and that there is no valid justification to output a different quasi-ultrametric. 

\subsection{Existence and uniqueness of admissible quasi-clustering methods: directed single linkage}\label{sec_existance_uniqueness_quasi_clustering}

We call a quasi-clustering method $\tilde{\ccalH}$ admissible if it satisfies axioms (\~A1) and (\~A2) and we want to find methods that are admissible with respect to these axioms. This is not difficult. Define the directed minimum chain cost $\tdu^*_X(x, x')$ between nodes $x$ and $x'$ as the minimum chain cost among all chains connecting $x$ to $x'$. Formally, for all $x, x' \in X$,
\begin{align}\label{eqn_nonreciprocal_chains} 
   \tdu^*_X(x, x') = \min_{C(x,x')} \,\,
                        \max_{i | x_i\in C(x,x')} A_X(x_i,x_{i+1}).
\end{align} 
Define the \emph{directed single linkage} (DSL) hierarchical quasi-clustering method $\tilde{\ccalH}^*$ as the one with output quasi-ultrametrics $(X, \tdu_X^*)=\tilde{\ccalH}^*(X, A_X)$ given by the directed minimum chain cost function $\tdu^*_X$.  The DSL method is valid and admissible as we show in the following proposition.

\begin{proposition}\label{prop_directed_axioms}
The hierarchical quasi-clustering method $\tilde{\ccalH}^*$ is valid and admissible. I.e., $\tdu^*_X$ defined by  \eqref{eqn_nonreciprocal_chains} is a quasi-ultrametric and $\tilde{\ccalH}^*$ satisfies axioms (\~A1)-(\~A2).
\end{proposition}

We next ask which other methods satisfy (\~A1)-(\~A2) and what special properties DSL has. As it turns out, DSL is the unique quasi-clustering method that is admissible with respect to (\~A1)-(\~A2) as we assert in the following theorem.

\begin{theorem}\label{theo_uniqueness_quasi_clustering}
Let $\tilde{\ccalH}$ be a valid hierarchical quasi-clustering method satisfying axioms (\~A1) and (\~A2). Then, $\tilde{\ccalH} \equiv \tilde{\ccalH}^*$ where $\tilde{\ccalH}^*$ is the DSL method with output quasi-ultrametrics as in \eqref{eqn_nonreciprocal_chains}.
\end{theorem}

In \cite{clust-um}, it was shown that single linkage is the only admissible hierarchical clustering method for finite metric spaces. Admissibility was defined by three axioms, two of which are undirected versions of (\~A1) and (\~A2). In \cite{Carlssonetal13}, they show that when replacing metric spaces by more general asymmetric networks, the uniqueness result is lost and an infinite number of methods satisfy the admissibility axioms. In our paper, by considering the more general framework of quasi-clustering methods, we recover the uniqueness result even for asymmetric networks. Moreover, Theorem \ref{theo_uniqueness_quasi_clustering} shows that the only admissible method is a directed version of single linkage. In this way, it becomes clear that the non-uniqueness result for asymmetric networks in \cite{Carlssonetal13} is originated in the symmetry mismatch between the input asymmetric network and the output symmetric dendrogram. When we allow the more general asymmetric quasi-dendrogram as output, the uniqueness result is recovered. 

DSL was identified as a natural extension of single linkage hierarchical clustering to asymmetric networks in \cite{boyd-asymmetric}. In our paper, by developing a framework to study hierarchical quasi-clustering methods and leveraging the equivalence result in Theorem \ref{theo_equivalence_quasi_dendrogram_quasi_ultrametric}, we show that DSL is the \emph{unique} admissible way of quasi-clustering asymmetric networks. Furthermore, stability and invariance properties are established in the following section.

\begin{remark}[Axiomatic strength and directed chaining effect]
\normalfont DSL, having a strong resemblance to single linkage hierarchical clustering on finite metric spaces, is likely to be sensitive to a directed version of the so called chaining effect \cite{clusteringref}. By requiring a weaker version of (\~A2), the most stringent of our two axioms, the uniqueness result in Theorem \ref{theo_uniqueness_quasi_clustering} is lost and density aware methods, that do not suffer from the chaining effect, become admissible. This direction, shown to be successful for finite metric spaces \cite{CarlssonMemoli10}, appears to be an interesting research avenue.
\end{remark}

\subsection{Stability and invariance properties of DSL} \label{sec:stab}

DSL is stable in the sense that if it is applied to similar networks then it outputs similar quasi-dendrograms. This notion has been used to study stability of clustering methods for finite metric spaces \cite{clust-um}. In order to formalize this concept, we define a notion of distance between networks. We define an analogue to the Gromov-Hausdorff distance \cite{book-gromov} between metric spaces, which we denote $d_{\mathcal{N}}$ and defines a legitimate metric on $\mathcal{N}$ (see \ref{sec_metric_on_networks} in supplementary material for details).  Since we may regard DSL as a map $\mathcal{N}\longrightarrow\tilde{\mathcal{U}}$ and $\tilde{\mathcal{U}}$ is a subset of $\mathcal{N}$, we are in a position in which we can use $d_{\mathcal{N}}$ to express the stability of  $\tilde{\ccalH}^*$.
 \begin{theorem}\label{thm:stab-dsl}
 For all $N_X,N_Y\in\mathcal{N},$ 
 $$d_{\mathcal{N}}\big( \tilde{\ccalH}^*(N_X), \tilde{\ccalH}^*(N_Y)\big)\leq d_{\mathcal{N}}(N_X,N_Y).$$
 \end{theorem}
Theorem \ref{thm:stab-dsl} states that the distance between the output quasi-ultrametrics is upper bounded by the distance between the input networks. Thus, for DSL, nearby networks yield nearby quasi-ultrametrics. This is important when we consider noisy dissimilarity data. Theorem \ref{thm:stab-dsl} ensures that noise has limited effect on output quasi-dendrograms. Furthermore, the theorem implies that DSL is permutation invariant; see \ref{sec_inv_supp} in supplementary material.

For a non-decreasing function $\psi:[0,\infty)\rightarrow [0,\infty)$ such that $\psi(a)=0$ if and only if $a=0$, and $N_X=(X,A_X)\in\mathcal{N}$ we write $\psi(N_X)$ to denote the network $(X,\psi(A_X))$. Any such $\psi$ will be referred to as a \emph{change of scale function}. Then, DSL is a scale invariant method as the following proposition asserts.
\begin{proposition}\label{prop:scale}
For all $N_X\in\mathcal{N}$ and all change of scale functions $\psi$ one has $\psi\big(\tilde{\ccalH}^*(N_X)\big)=\tilde{\ccalH}^*\big(\psi(N_X)\big)$.
\end{proposition}
Since Proposition \ref{prop:scale} asserts that the quasi-ultrametric outcome is transformed by the same function $\psi$ that alters the dissimilarity function in the original network, DSL is invariant to change of units. More precisely, in terms of quasi-dendrograms, a transformation of dissimilarities through $\psi$ results in a transformed quasi-dendrogram where the order in which influences between nodes arise is the same as in the original one while the resolution at which they appear changes according to $\psi$. For further invariances of DSL, see \ref{sec_inv_supp} in the supplementary materials. 


\subsection{Algorithms}\label{sec_algorithms}

In this section we interpret $A_X$ as a matrix of dissimilarities and $\tdu^*_X$ as a symmetric matrix with entries corresponding to the quasi-ultrametric values $\tdu^*_X(x,x')$ for all $x, x' \in X$. By \eqref{eqn_nonreciprocal_chains}, DSL quasi-clustering searches for directed chains of minimum infinity norm cost in $A_X$ to construct the matrix $\tdu^*_X$. This operation can be performed algorithmically using matrix powers in the dioid algebra $(\reals^+\cup\{+\infty\},\min,\max)$ \cite{GondranMinoux08}.

In the dioid algebra $(\reals^+\cup\{+\infty\},\min,\max)$ the regular sum is replaced by the minimization operator and the regular product by maximization. Using $\oplus$ and $\otimes$ to denote sum and product on this dioid algebra we have $a\oplus b := \min(a,b)$ and $a\otimes b := \max(a,b)$ for all $a, b \in \reals^+\cup\{+\infty\}$. The matrix product $A\otimes B$ is therefore given by the matrix with entries
\begin{equation}\label{def_star}
   \big[A \otimes B\big]_{ij}  
       \ =\ \bigoplus_{k=1}^n \big(A_{ik} \otimes B_{kj} \big) 
       \ =\ \min_{k\in[1,n]} \, \max \big(A_{ik},B_{kj} \big).
\end{equation}
Dioid powers $A_X^{(k)}:=A_X\otimes A_X^{(k-1)}$ with $A_X^{(1)}=A_X$ of a dissimilarity matrix are related to quasi-ultrametric matrices $\tdu$. For instance, the elements of the dioid power $\tdu^{(2)}$ of a given quasi-ultrametric matrix $\tdu$ are given by
\begin{equation}\label{def_diod_algebra_ultrametric}
   \big[\tdu^{(2)}\big]_{ij}  
       = \min_{k\in[1,n]} \, \max \big(\tdu_{ik},\tdu_{kj} \big).
\end{equation}
Since $\tdu$ satisfies the strong triangle inequality we have that $\tdu_{ij}\leq\max (\tdu_{ik},\tdu_{kj})$ for all $k$. And for $k=j$ in particular we further have that $\max (\tdu_{ik},\tdu_{kj})=\max (\tdu_{ij},\tdu_{jj})=\max(\tdu_{ij},0)=\tdu_{ij}$. Combining these two observations it follows that the result of the minimization in \eqref{def_diod_algebra_ultrametric} is $[\tdu^{(2)}]_{ij} =  \tdu_{ij}$ since none of its arguments is smaller that $\tdu_{ij}$ and one of them is exactly $\tdu_{ij}$. This being valid for all $i,j$ implies $\tdu^{(2)} =  \tdu$.
Furthermore, a matrix satisfying $\tdu^{(2)} =  \tdu$ is such that $\tdu_{ij} = [\tdu^{(2)}]_{ij} = \min_{k\in[1,n]} \, \max (\tdu_{ik}, \tdu_{kj} ) \leq \max(\tdu_{ik},\tdu_{kj} )$ for all $k$, which is just a restatement of the strong triangle inequality. Therefore, a non-negative matrix $\tdu$ represents a finite quasi-ultrametric space if and only if $\tdu^{(2)} =  \tdu$ and only the diagonal elements are null. Building on this fact, we state the following algorithm to compute the quasi-ultrametric output by the DSL method.

\begin{proposition}\label{prop_algorithms}
For every network $(X, A_X)$ with $|X|=n$, the quasi-ultrametric $\tdu_X^*$ is given by
\begin{equation}\label{eqn_algo_quasi_ultrametric}
\tdu_X^*=A_X^{(n-1)},
\end{equation}
where the operation $(\cdot)^{(n-1)}$ denotes the $(n-1)$st matrix power in the dioid algebra $(\reals^+\cup\{+\infty\},\min,\max)$ with matrix product as defined in \eqref{def_star}.
\end{proposition}

Matrix powers in dioid algebras are tractable operations. Indeed, there exist sub cubic dioid power algorithms \cite{VassilevskaEtal09, DuanPettie09} of complexity $\mathrm{O}(n^{2.688})$. Thus, Proposition \ref{prop_algorithms} shows computational tractability of the DSL quasi-clustering method. There exist related methods with lower complexity. For instance, Tarjan's method \cite{tarjan-improved}, which takes as input an asymmetric network but in contrast to our method enforces symmetry in its output, runs in time $\mathrm{O}(n^2\log n)$ for complete networks. It seems of interest to ascertain whether one might be able to modify his algorithm to suit our (asymmetric) output construction. In the following section we use \eqref{eqn_algo_quasi_ultrametric} to quasi-cluster a real-world network.


\section{Applications}\label{sec_applications}

The number of migrants from state to state is published yearly by the geographical mobility section of the U.S. census bureau \cite{USmigration}. We denote as $S$ the set containing every state plus the District of Columbia and as $A_S: S \times S \to \reals_+$ a migrational dissimilarity such that $A_S(s, s)=0$ for all $s \in S$ and $A_S(s, s')$ for all $s \neq s' \in S$ is a monotonically decreasing function of the fraction of immigrants to state $s'$ that come from $s$ (see \ref{sec_sup_mat_applications} in supplementary material for details). A small dissimilarity from state $s$ to state $s'$ implies that, among all the immigrants into $s'$, a high percentage comes from $s$. We then construct the asymmetric network $N_S=(S, A_S)$ with node set $S$ and dissimilarities $A_S$. The application of hierarchical clustering to migration data has been extensively investigated by Slater, see \cite{slater1976hierarchical,slater1984partial}.

The outcome of applying DSL with output quasi-ultrametric defined in \eqref{eqn_nonreciprocal_chains} to the migration network $N_S$ is computed via \eqref{eqn_algo_quasi_ultrametric}. By Theorem \ref{theo_equivalence_quasi_dendrogram_quasi_ultrametric}, the output quasi-ultrametric is equivalent to a quasi-dendrogram $\tilde{D}^*_S= (D^*_S, E^*_S)$. By analyzing the dendrogram component $D^*_S$ of the quasi-dendrogram $\tilde{D}^*_S$, the influence of geographical proximity in migrational preference is evident; see Fig. \ref{fig_nonreciprocal_us_dendrogram} in Section \ref{sec_sup_mat_applications} of the supplementary material.

\begin{figure}[t]
\centering
\input{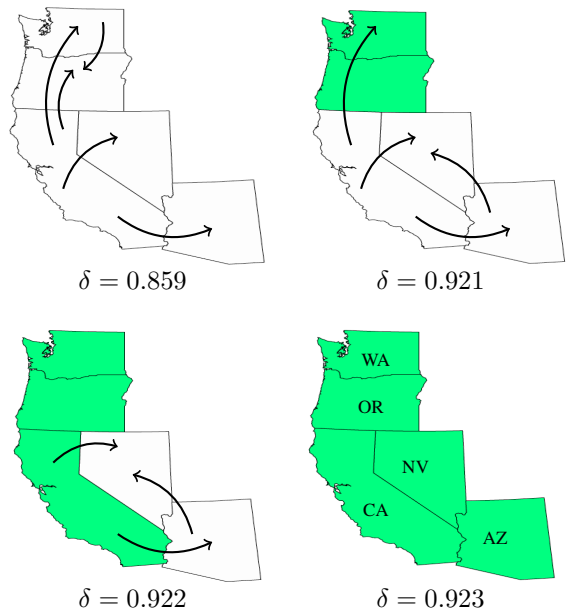}
\caption{Directed single linkage quasi-clustering method applied to the extended West Coast migration flow.}
\vspace{-0.2in}
\label{fig_quasi-clustering_west_coast}
\end{figure}

To facilitate display and understanding, we do not present quasi-partitions for all the nodes and resolutions. Instead, we restrict the quasi-ultrametric to a subset of states representing an extended West Coast including Arizona and Nevada. In Fig. \ref{fig_quasi-clustering_west_coast}, we depict quasi-partitions at four relevant resolutions of the quasi-dendrogram equivalent to the restricted quasi-ultrametric.
States represented with the same color in the maps in Fig. \ref{fig_quasi-clustering_west_coast} are part of the same cluster at the given resolution and states in white form singleton clusters. Arrows between clusters for a given resolution $\delta$ represent the edge set $E^*_S(\delta)$ which we interpret as a migrational influence relation between the blocks of states.

The DSL quasi-clustering method $\tilde{\ccalH}^*$ captures not only the formation of clusters but also the asymmetric influence between them. E.g. the quasi-partition in Fig. \ref{fig_quasi-clustering_west_coast} for resolution $\delta=0.859$ is of little interest since every state forms a singleton cluster. The influence structure, however, reveals a highly asymmetric migration pattern. At this resolution California has migrational influence over every other state in the region as depicted by the four arrows leaving California and entering each of the other states. This influence can be explained by the fact that California contains the largest urban areas of the region such as Los Angeles. Hence, these urban areas attract immigrants from all over the country, reducing the proportional immigration into California from its neighbors and generating the asymmetric influence structure observed. Since this influence structure defines a partial order over the clusters, the quasi-partition at resolution $\delta=0.859$ permits asserting the reasonable fact that California is the dominant migration force in the region.

At larger resolutions we can ascertain the relative importance of clusters. At resolution $\delta=0.921$ we can say that California is more important than the cluster formed by Oregon and Washington as well as more important than Arizona and Nevada. We can also see that Arizona precedes Nevada in the migration ordering at this resolution while the remaining pairs of the ordering are undefined. At resolution $\delta=0.922$ there is an interesting pattern as we can see the cluster formed by the three West Coast states preceding Arizona and Nevada in the partial order. At this resolution the partial order also happens to be a total order as Arizona is seen to precede Nevada. This is not true in general as we have already seen.

Hierarchical quasi-clustering methods can also be used to study, e.g., the relations between sectors of an economy. Due to space restrictions, we include this second application in \ref{sec_sup_mat_applications} in the supplementary material.

\section{Conclusion}\label{sec_conclusion}

When clustering asymmetric networks, requiring the output to be symmetric -- as in hierarchical clustering -- might be undesirable. Hence, we defined quasi-dendrograms, a generalization of dendrograms that admits asymmetric relations, and developed a theory for quasi-clustering methods. We formalized the notion of admissibility by introducing two axioms. Under this framework, we showed that DSL is the unique admissible method. We pointed out that less stringent frameworks that give rise to new admissible methods can be explored by weakening the Directed Axiom of Transformation. Furthermore, we proved an equivalence between quasi-dendrograms and quasi-ultrametrics that generalizes the well-known equivalence between dendrograms and ultrametrics, and established the stability and invariance properties of the DSL method. Finally, we illustrated the application of DSL to a migration network.

\section*{Acknowledgments} 
 
Work in this paper is supported by NSF CCF-0952867, AFOSR MURI FA9550-10-1-0567, DARPA GRAPHS FA9550-12-1-0416, AFOSR FA9550-09-0-1-0531, AFOSR FA9550-09-1-0643, NSF DMS 0905823, and NSF DMS-0406992.

\newpage

\bibliography{clustering_biblio}
\bibliographystyle{icml2014}

\newpage
\begin{appendices}

\section{Supplementary Material}
\subsection{Proof of Theorem \ref{theo_equivalence_quasi_dendrogram_quasi_ultrametric}}

In order to show that $\Psi$ is a well-defined map, we must show that $\Psi(\tdD_X)$ is a quasi-ultrametric network for every quasi-dendrogram $\tdD_X$. Given an arbitrary quasi-dendrogram $\tdD_X=(D_X, E_X)$, for a particular $\delta' \geq 0$ consider the quasi-partition $\tdD_X(\delta')$. Consider the range of resolutions $\delta$ associated with such quasi-partition. I.e.,
\begin{equation}\label{eqn_theo_pf_dendrograms_as_quasi_ultrametrics_00}
\{\delta \geq 0 \given \tdD_X(\delta)=\tdD_X(\delta')\}.
\end{equation}
Right continuity (\~D4) of $\tdD_X$ ensures that the minimum of the set in \eqref{eqn_theo_pf_dendrograms_as_quasi_ultrametrics_00} is well-defined and hence definition \eqref{eqn_theo_dendrograms_as_quasi_ultrametrics_10} is valid. To prove that $\tdu_X$ in \eqref{eqn_theo_dendrograms_as_quasi_ultrametrics_10} is a quasi-ultrametric we need to show that it attains non-negative values as well as the identity and strong triangle inequality properties. That $\tdu_X$ attains non-negative values is clear from the definition \eqref{eqn_theo_dendrograms_as_quasi_ultrametrics_10}. The identity property is implied by the first boundary condition in (\~D1). Since $[x]_0=[x]_0$ for all $x \in X$, we must have $\tdu_X(x, x)=0$. Conversely, since for all $x \neq x' \in X$, $([x]_0, [x']_0) \not\in E_X(0)$ and $[x]_0 \neq [x']_0$ we must have that $\tdu_X(x, x')>0$ for $x \neq x'$ and the identity property is satisfied.
To see that $\tdu_X$ satisfies the strong triangle inequality in \eqref{eqn_def_strong_triangle_inequality}, consider nodes $x$, $x'$, and $x''$ such that the lowest resolution for which $[x]_\delta = [x'']_\delta$ or $([x]_\delta, [x'']_\delta) \in E_X(\delta)$ is $\delta_1$ and the lowest resolution for which $[x'']_\delta = [x']_\delta$ or $([x'']_\delta, [x']_\delta) \in E_X(\delta)$ is $\delta_2$. Right continuity (\~D4) ensures that these lowest resolutions are well-defined. According to \eqref{eqn_theo_dendrograms_as_quasi_ultrametrics_10} we then have
\begin{alignat}{3}\label{eqn_theo_pf_dendrograms_as_quasi_ultrametrics_10}
   \tdu_X(x, x'') = \delta_1, \nonumber\\
   \tdu_X(x'',x') = \delta_2.
\end{alignat}
Denote by $\delta_0:=\max(\delta_1,\delta_2)$. From the equivalence hierarchy (\~D2) and influence hierarchy (\~D3) properties, it follows that $[x]_{\delta_0}=[x'']_{\delta_0}$ or $([x]_{\delta_0}, [x'']_{\delta_0}) \in E_X({\delta_0})$ and $[x'']_{\delta_0}=[x']_{\delta_0}$ or $([x'']_{\delta_0}, [x']_{\delta_0}) \in E_X({\delta_0})$. Furthermore, from transitivity (QP2) of the quasi-partition $\tdD_X(\delta_0)$, it follows that $[x]_{\delta_0}=[x']_{\delta_0}$ or $([x]_{\delta_0}, [x']_{\delta_0}) \in E_X({\delta_0})$. Using the definition in \eqref{eqn_theo_dendrograms_as_quasi_ultrametrics_10} for $x$, $x'$ we conclude that
\begin{equation}\label{eqn_theo_pf_dendrograms_as_quasi_ultrametrics_20}
   \tdu_X(x,x') \leq \delta_0.
\end{equation}
By definition $\delta_0:=\max(\delta_1,\delta_2)$, hence we substitute this expression in \eqref{eqn_theo_pf_dendrograms_as_quasi_ultrametrics_20} and compare with \eqref{eqn_theo_pf_dendrograms_as_quasi_ultrametrics_10} to obtain
\begin{equation}\label{eqn_theo_pf_dendrograms_as_ultrametrics_30}
   \tdu_X(x,x') \! \leq \! \max(\delta_1,\delta_2) \! = \! \max \Big(\tdu_X(x,x''), \tdu_X(x'',x')\Big).
\end{equation}
Consequently, $\tdu_X$ satisfies the strong triangle inequality and is therefore a quasi-ultrametric, proving that the map $\Psi$ is well-defined.

For the converse result, we need to show that $\Upsilon$ is a well-defined map. Given a quasi-ultrametric $\tdu_X$ on a node set $X$ and a resolution $\delta \geq 0$, we first define the relation
\begin{equation}\label{eqn_theo_pf_dendrograms_as_ultrametrics_31}
x \leadsto_{\tdu_X(\delta)} x' \quad \iff \quad \tdu_X(x, x') \leq \delta,
\end{equation}
for all $x, x' \in X$. Notice that $ \leadsto_{\tdu_X(\delta)}$ is a quasi-equivalence relation as defined in Definition \ref{def_quasi_equivalence} for all $\delta \geq 0$. The reflexivity property is implied by the identity property of the quasi-ultrametric $\tdu_X$ and transitivity is implied by the fact that $\tdu_X$ satisfies the strong triangle inequality. Furthermore, definitions \eqref{eqn_theo_dendrograms_as_quasi_ultrametrics_20} and \eqref{eqn_theo_dendrograms_as_quasi_ultrametrics_30} are just reformulations of \eqref{eqn_quasi_equiv_equiv} and \eqref{eqn_quasi_equiv_edges_quasi_partition} respectively, for the special case of the quasi-equivalence defined in \eqref{eqn_theo_pf_dendrograms_as_ultrametrics_31}. Hence, Proposition \ref{prop_quasi_equiv_quasi_part} guarantees that $\Upsilon(X, \tdu_X)=\tdD_X(\delta)=(D_X(\delta), E_X(\delta))$ is a quasi-partition for every resolution $\delta \geq 0$. In order to show that $\Upsilon$ is well-defined, we need to show that these quasi-partitions are nested, i.e. that $\tdD_X$ satisfies (\~D1)-(\~D4).

The first boundary condition in (\~D1) is implied by \eqref{eqn_theo_dendrograms_as_quasi_ultrametrics_20} and the identity property of $\tdu_X$. The second boundary condition in (\~D1) is implied by the fact that $\tdu_X$ takes finite real values on a finite domain since the node set $X$ is finite. Hence, any $\delta_0$ satisfying
\begin{equation}\label{eqn_theo_pf_dendrograms_as_ultrametrics_32}
\delta_0 \geq \max_{x, x' \in X} \tdu_X(x, x'),
\end{equation}
is a valid candidate to show fulfillment of (\~D1).

To see that $\tdD_X$ satisfies (\~D2) assume that for a resolution $\delta_1$ we have two nodes $x, x' \in X$ such that $x \sim_{\tdu_X(\delta_1)} x'$ as in \eqref{eqn_theo_dendrograms_as_quasi_ultrametrics_20}, then it follows that $\max \big( \tdu_X(x,x'), \tdu_X(x',x) \big) \leq \delta_1$. Thus, if we pick any $\delta_2 > \delta_1$ it is immediate that $\max \big( \tdu_X(x,x'), \tdu_X(x',x) \big) \leq \delta_2$ which by \eqref{eqn_theo_dendrograms_as_quasi_ultrametrics_20} implies that $x \sim_{\tdu_X(\delta_2)} x'$.

Fulfillment of (\~D3) can be shown in a similar way as fulfillment of (\~D2). Given a scalar $\delta_1 \geq 0$ and $x, x' \in X$, if $([x]_{\delta_1}, [x']_{\delta_1}) \in E_X(\delta_1)$ then by \eqref{eqn_theo_dendrograms_as_quasi_ultrametrics_30} we have that
\begin{equation}\label{eqn_theo_dendrograms_as_quasi_ultrametrics_33}
   \min_{x_1 \in [x]_{\delta_1},x_2 \in [x']_{\delta_1}} \tdu_X(x_1, x_2) \leq \delta_1.
\end{equation}
From property (\~D2), we know that for all $x \in X$, $[x]_{\delta_1} \subset [x]_{\delta_2}$ for all $\delta_2 > \delta_1$. Hence, two things might happen. Either $\max(\tdu_X(x, x'), \tdu_X(x', x)) \leq \delta_2$ in which case $[x]_{\delta_2}=[x']_{\delta_2}$ or it might be that $[x]_{\delta_2} \neq [x']_{\delta_2}$ but 
\begin{equation}\label{eqn_theo_dendrograms_as_quasi_ultrametrics_34}
   \min_{x_1 \in [x]_{\delta_2},x_2 \in [x']_{\delta_2}} \tdu_X(x_1, x_2) \leq \delta_1 < \delta_2,
\end{equation}
which implies that $([x]_{\delta_2}, [x']_{\delta_2}) \in E_X(\delta_2)$. Consequently, (\~D3) is satisfied.

Finally, to see that $\tdD_X$ satisfies the right continuity condition (\~D4), for each $\delta \geq 0$ such that $\tdD_X(\delta) \neq ( \{ X\}, \emptyset )$ we may define $\epsilon(\delta)$ as any positive scalar satisfying
\begin{equation}\label{eqn_epsilon_d_3_ultrametric_quasi_dendrogram}
0 < \epsilon(\delta) < \displaystyle \min_{\substack{x, x' \in X \\ \text{s.t.} \,\, \tdu_X(x, x') > \delta}} \tdu_X(x, x')- \delta,
\end{equation} 
where the finiteness of $X$ ensures that $\epsilon(\delta)$ is well-defined.
Hence, \eqref{eqn_theo_dendrograms_as_quasi_ultrametrics_20} and \eqref{eqn_theo_dendrograms_as_quasi_ultrametrics_30} guarantee that $\tdD_X(\delta)=\tdD_X(\delta')$ for $\delta' \in [\delta, \delta + \epsilon(\delta)]$. For all other resolutions $\delta$ such that $\tdD_X(\delta) = ( \{ X\}, \emptyset)$, right continuity is trivially satisfied since the quasi-dendrogram remains unchanged for increasing resolutions. Consequently, $\Upsilon(X, \tdu_X)$ is a valid quasi-dendrogram for every quasi-ultrametric network $(X, \tdu_X)$, proving that $\Upsilon$ is well-defined.

In order to conclude the proof, we need to show that $\Psi \circ \Upsilon$ and $\Upsilon \circ \Psi$ are the identities on $\tilde{\mathcal{U}}$ and $\tilde{\mathcal{D}}$, respectively. To see why the former is true, pick any quasi-ultrametric network $(X, \tdu_X)$ and consider an arbitrary pair of nodes $x, x' \in X$ such that $\tdu_X(x, x')=\delta_0$. Also, consider the ultrametric network $\Psi \circ \Upsilon (X, \tdu_X):=(X, \tdu^*_X)$. From \eqref{eqn_theo_dendrograms_as_quasi_ultrametrics_20} and \eqref{eqn_theo_dendrograms_as_quasi_ultrametrics_30}, in the quasi-dendrogram $\Upsilon (X, \tdu_X)$ there is no influence from $x$ to $x'$ for resolutions $\delta < \delta_0$ and at resolution $\delta=\delta_0$ either an edge appears from $[x]_{\delta_0}$ to $[x']_{\delta_0}$, or both nodes merge into one single cluster. In any case, when we apply $\Psi$ to the resulting quasi-dendrogram, we obtain $\tdu^*_X(x, x')=\delta_0$. Since $x, x' \in X$ were chosen arbitrarily, we have that $\tdu_X = \tdu^*_X$, showing that $\Psi \circ \Upsilon$ is the identity on $\tilde{\mathcal{U}}$. A similar argument shows that $\Upsilon \circ \Psi$ is the identity on $\tilde{\mathcal{D}}$.


\subsection{Proof of Proposition \ref{prop_directed_axioms}}
For this proof, we introduce the concept of chain concatenation. Given two chains $C(x, x')=[x=x_0, x_1, ... , x_l=x']$ and $C(x', x'')=[x'=x'_0, x'_1, ... , x'_{l'}=x'']$ such that the end point $x'$ of the first one coincides with the starting point of the second one, define the \emph{concatenated chain} $C(x, x') \uplus C(x',x'')$ as
\begin{align}\label{eqn_definition_concatenation}
   &C(x, x') \uplus C(x',x'')  \nonumber \\ &\qquad\quad
      := [x=x_0,\ldots , x_l=x'=x'_0,\ldots , x'_{l'}=x'']. 
\end{align}
For the method $\tilde{\ccalH}^{*}$ to be a properly defined hierarchical quasi-clustering method, we need to establish that $\tdu^{*}_X$ is a valid ultrametric. To see that $\tdu^{*}_X(x,x')=0$ if and only if $x=x'$, notice that when $x=x'$, the chain $C(x, x)=[x, x]$ has null cost and when $x \neq x'$ any chain must contain at least one link with strictly positive cost. To verify that the strong triangle inequality in \eqref{eqn_def_strong_triangle_inequality} holds, let $C^*(x,x'')$ and $C^*(x'',x')$ be chains that achieve the minimum in \eqref{eqn_nonreciprocal_chains} for $\tdu^{*}_X(x,x'')$ and $\tdu^{*}_X(x'',x')$, respectively. The maximum cost in the concatenated chain $C(x,x')=C^*(x,x'') \uplus C^*(x'',x')$ does not exceed the maximum cost in each individual chain. Thus, while the maximum cost may be smaller on a different chain, the chain $C(x,x')$ suffices to bound $\tdu^{*}_X(x,x') \leq \max \big( \tdu^{*}_X(x,x''), \tdu^{*}_X(x'',x')\big)$ as in \eqref{eqn_def_strong_triangle_inequality}. 

To show fulfillment of Axiom (\~A1), pick an arbitrary two-node network $\vec{\Delta}_2(\alpha, \beta):= (\{p,q\}, A_{p,q})$ with $A_{p,q}(p,q)=\alpha$ and $A_{p,q}(q,p)=\beta$ for some $\alpha, \beta >0$ and denote by $(\{p,q\}, \tdu^*_{p,q})=\tilde{\ccalH}^*(\vec{\Delta}_2(\alpha, \beta))$. Then, we have $\tdu^*_{p,q}(p,q)=\alpha$ and $\tdu^*_{p,q}(q,p)=\beta$ because there is only one possible chain selection in each direction [cf. \eqref{eqn_nonreciprocal_chains}]. To prove that Axiom (\~A2) is satisfied consider arbitrary points $x,x' \in X$ and denote by $C^*(x, x')$ one chain achieving the minimum chain cost in \eqref{eqn_nonreciprocal_chains},
\begin{align}\label{eqn_theo_nonreciprocal_axioms_pf_30} 
   \tdu^*_X(x, x') = \max_{i | x_i\in C^*(x,x')} A(x_i,x_{i+1}).
\end{align} 

Consider the transformed chain $C_Y(\phi(x),\phi(x'))=[\phi(x)=\phi(x_0),\ldots, \phi(x_l)=\phi(x')]$ in the space $Y$. Since the map $\phi:X\to Y$ reduces dissimilarities we have that for all links in this chain $A_Y(\phi(x_i),\phi(x_{i+1}))\leq A_X(x_i,x_{i+1})$. Consequently,
\begin{align}\label{eqn_theo_nonreciprocal_axioms_pf_40}
    &\max_{i | \phi(x_i)\in C_Y(\phi(x),\phi(x'))} A_Y(\phi(x_i),\phi(x_{i+1})) 
                       \\\nonumber &\hspace{40mm}
           \leq \max_{i | x_i\in C^*(x,x')} A_X(x_i,x_{i+1}).
\end{align}
Further note that the minimum chain cost $\tdu^*_Y(\phi(x), \phi(x'))$ among all chains linking $\phi(x)$ to $\phi(x')$ cannot exceed the cost in the given chain $C_Y(\phi(x),\phi(x'))$. Combining this observation with the inequality in \eqref{eqn_theo_nonreciprocal_axioms_pf_40} it follows that
\begin{align}\label{eqn_theo_nonreciprocal_axioms_pf_50} 
   \tdu^*_Y(\phi(x),\phi(x')) \! \leq \! \max_{i | x_i\in C^*(x,x')} A_X(x_i,x_{i+1})
                                \!  = \! \tdu^*_X(x, x'),
\end{align} 
where we also used \eqref{eqn_theo_nonreciprocal_axioms_pf_30} to write the equality. Expression \eqref{eqn_theo_nonreciprocal_axioms_pf_50} ensures fulfillment of Axiom (\~A2), as wanted.

\subsection{Proof of Theorem \ref{theo_uniqueness_quasi_clustering}}
In proving this theorem, the concept of \emph{separation} of a network is useful. Given an arbitrary network $(X, A_X)$, its separation $\sep(X, A_X)$ is defined as the minimum positive dissimilarity in the network, that is
\begin{equation}\label{eqn_def_separation_network}
\sep(X, A_X) = \min_{x \neq x'} A_X(x, x').
\end{equation}
The following auxiliary result is useful in showing Theorem \ref{theo_uniqueness_quasi_clustering}.
\begin{lemma}\label{lemma_axiom_redundancy}
A network $N=(X, A_X)$ and a positive constant $\delta$ are given. Then, for any pair of nodes $x,x' \in X$ whose minimum chain cost [cf. \eqref{eqn_nonreciprocal_chains}] satisfies
\begin{align}\label{eqn_lem_axiom_redundancy_01} 
   \tdu^*_X(x, x') \geq \delta,
\end{align} 
there exists a partition $P_\delta(x,x')=\{B_\delta(x), B_\delta(x')\}$ of the node space $X$ into blocks $B_\delta(x)$ and $B_\delta(x')$ with $x \in B_\delta(x)$ and $x' \in B_\delta(x')$ such that for all points $b \in B_\delta(x)$ and $b' \in B_\delta(x')$
\begin{align}\label{eqn_lem_axiom_redundancy_02} 
   A_X(b, b') \geq \delta.
\end{align} \end{lemma}
\begin{myproof}
We prove this result by contradiction. If a partition $P_\delta(x,x')=\{B_\delta(x), B_\delta(x')\}$ with $x \in B_\delta(x)$ and $x' \in B_\delta(x)$ and satisfying \eqref{eqn_lem_axiom_redundancy_02} does not exist for all pairs of points $x,x'\in X$ satisfying \eqref{eqn_lem_axiom_redundancy_01}, then there is at least one pair of nodes $x,x'\in X$ satisfying \eqref{eqn_lem_axiom_redundancy_01} such that for {\it all} partitions of $X$ into two blocks $P=\{B, B'\}$ with $x \in B$ and $x' \in B'$ we can find at least a pair of elements $b_P \in B$ and $b'_P \in B'$ for which
\begin{equation}\label{eqn_lem_axiom_redundancy_03} 
   A_X(b_P, b'_P) < \delta.
\end{equation}
Begin by considering the partition $P_1=\{B_1, B'_1\}$ where $B_1 = \{ x \}$ and $B'_1 = X \backslash \{x\}$. Since \eqref{eqn_lem_axiom_redundancy_03} is true for all partitions having $x\in B$ and $x'\in B'$ and $x$ is the unique element of $B_1$, there must exist a node $b'_{P_1} \in B'_1$ such that  
\begin{equation}\label{eqn_lem_axiom_redundancy_04} 
   A_X(x, b'_{P_1}) <  \delta. 
\end{equation}
Hence, the chain $C(x, b'_{P_1})= [x, b'_{P_1}]$ composed of these two nodes has cost smaller than $\delta$. Moreover, since $\tdu^*_X(x, b'_{P_1})$ represents the minimum cost among all chains $C(x, b'_{P_1})$ linking $x$ to $b'_{P_1}$, we can assert that
\begin{equation}\label{eqn_lem_axiom_redundancy_04_1} 
   \tdu^*_X(x, b'_{P_1}) \leq  A_X(x, b'_{P_1}) <  \delta.
\end{equation}
Consider now the partition $P_2=\{B_2, B'_2\}$ where $B_2= \{ x, b'_{P_1} \}$ and $B'_2=X \backslash B_2$. From \eqref{eqn_lem_axiom_redundancy_03}, there must exist a node $b'_{P_2} \in B'_2$ that satisfies at least one of the two following conditions 
\begin{align}
   &A_X(x, b'_{P_2}) <  \delta,  \label{eqn_lem_axiom_redundancy_05} \\
   &A_X(b'_{P_1}, b'_{P_2}) <  \delta. \label{eqn_lem_axiom_redundancy_06}
\end{align}
If \eqref{eqn_lem_axiom_redundancy_05} is true, the chain $C(x, b'_{P_2})=[x, b'_{P_2}]$ has cost smaller than $\delta$. If \eqref{eqn_lem_axiom_redundancy_06} is true, we combine the dissimilarity bound  with the one in \eqref{eqn_lem_axiom_redundancy_04} to conclude that the chain $C(x, b'_{P_2})=[x, b'_{P_1}, b'_{P_2}]$ has cost smaller than $\delta$. In either case we conclude that there exists a chain $C(x, b'_{P_2})$ linking $x$ to $b'_{P_2}$ whose cost is smaller than $\delta$. Therefore, the minimum chain cost must satisfy
\begin{equation}\label{eqn_lem_axiom_redundancy_04_2} 
\tdu^*_X(x, b'_{P_2}) <  \delta.
\end{equation}
Repeat the process by considering the partition $P_3$ with $B_3= \{ x, b'_{P_1}, b'_{P_2} \}$ and $B'_3=X\backslash B_3$. As we did in arguing \eqref{eqn_lem_axiom_redundancy_05}-\eqref{eqn_lem_axiom_redundancy_06} it must follow from \eqref{eqn_lem_axiom_redundancy_03} that there exists a point $b'_{P_3}$ such that at least one of the dissimilarities $A_X(x, b'_{P_3})$, $A_X(b'_{P_1}, b'_{P_3})$, or $A_X( b'_{P_2}, b'_{P_3})$ is smaller than $\delta$. This observation implies that at least one of the chains $[x, b'_{P_3}]$, $[x, b'_{P_1}, b'_{P_3}]$, $[x, b'_{P_2}, b'_{P_3}]$, or $[x, b'_{P_1}, b'_{P_2}, b'_{P_3}]$ has cost smaller than $\delta$ from where it follows that
\begin{equation}\label{eqn_lem_axiom_redundancy_04_3} 
   \tdu^*_X(x, b'_{P_3}) <  \delta.
\end{equation}
This recursive construction can be repeated $n-1$ times to obtain partitions $P_1, P_2, ... , P_{n-1}$ and corresponding nodes $b'_{P_1}, b'_{P_2}, ... b'_{P_{n-1}}$ such that the minimum chain cost satisfies
\begin{equation}\label{eqn_lem_axiom_redundancy_04_4} 
   \tdu^*_X(x, b'_{P_i}) <  \delta, \qquad \text{for all\ } i.
\end{equation}
Observe now that the nodes $b'_{P_i}$ are distinct by construction and distinct from $x$. Since there are $n$ nodes in the network it must be that $x'=b'_{P_k}$ for some $i \in \{1, \ldots , n-1\}$. It then follows from \eqref{eqn_lem_axiom_redundancy_04_4} that
\begin{equation}\label{eqn_lem_axiom_redundancy_04_5} 
\tdu^*_X(x, x') <  \delta.
\end{equation}
This is a contradiction because $x,x'\in X$ were assumed to satisfy \eqref{eqn_lem_axiom_redundancy_01}. Thus, the assumption that \eqref{eqn_lem_axiom_redundancy_03} is true for {\it all} partitions is incorrect. Hence, the claim that there exists a partition $P_\delta(x,x')=\{B_\delta(x), B_\delta(x')\}$ satisfying \eqref{eqn_lem_axiom_redundancy_02} must be true. \end{myproof}

Returning to the main proof, given an arbitrary network $N=(X, A_X)$ denote as $(X,\tdu_X)=\tilde{\ccalH}(X,A_X)$ the output quasi-ultrametric resulting from application of an arbitrary admissible quasi-clustering method $\tilde{\ccalH}$. We will show that for all $x, x' \in X$
\begin{equation}\label{eqn_inequality_unicity_directed}
  \tdu_X^*(x, x') \leq\tdu_X(x, x') \leq\tdu_X^*(x, x').
\end{equation}
To prove the rightmost inequality in \eqref{eqn_inequality_unicity_directed} we begin by showing that the dissimilarity function $A_X$ acts as an upper bound on all admissible quasi-ultrametrics $\tdu_X$, i.e.
\begin{equation}\label{eqn_dissimilarity_upper_bound_directed}
\tilde{u}_X(x, x') \leq A_X(x, x'),
\end{equation}
for all $x, x' \in X$. To see this, suppose $A_X(x, x')=\alpha$ and $A_X(x', x)=\beta$. Define the two-node network $N_{p,q}=(\{p,q\}, A_{p,q})$ where $A_{p,q}(p,q)=\alpha$ and $A_{p,q}(q,p)=\beta$ and denote by $(\{p, q\}, \tilde{u}_{p,q})=\tilde{\ccalH}(N_{p,q})$ the output of applying the method $\tilde{\ccalH}$ to the network $N_{p,q}$. From axiom (\~A1), we have $\tilde{\ccalH}(N_{p,q})=N_{p,q}$, in particular
\begin{equation}\label{eqn_dissimilarity_two_node_network}
\tilde{u}_{p,q}(p, q) = A_{p,q}(p, q)= A_{X}(x, x').
\end{equation}
Moreover, notice that the map $\phi:\{p,q\} \to X$, where $\phi(p)=x$ and $\phi(q)=x'$ is a dissimilarity reducing map, i.e. it does not increase any dissimilarity, from $N_{p,q}$ to $N$. Hence, from axiom (\~A2), we must have
\begin{equation}\label{eqn_quasi-ultra_two_node_network}
\tdu_{p,q}(p,q) \geq \tdu_{X}(\phi(p),\phi(q)) = \tdu_{X}(x,x').
\end{equation}
Substituting \eqref{eqn_dissimilarity_two_node_network} in \eqref{eqn_quasi-ultra_two_node_network}, we obtain \eqref{eqn_dissimilarity_upper_bound_directed}.

Consider now an arbitrary chain $C(x, x')=[x=x_0, x_1, \ldots , x_l=x']$ linking nodes $x$ and $x'$. Since $\tilde{u}_X$ is a valid quasi-ultrametric, it satisfies the strong triangle inequality \eqref{eqn_def_strong_triangle_inequality}. Thus, we have that
\begin{align}\label{eqn_stron_triangle_directed}
\tilde{u}_{X}(x, x') & \leq \max_{i|x_i \in C(x, x')} \tdu_{X}(x_i, x_{i+1}) \nonumber \\
& \leq \max_{i|x_i \in C(x, x')} A_X(x_i, x_{i+1}),
\end{align}
where the last inequality is implied by \eqref{eqn_dissimilarity_upper_bound_directed}. Since by definition $C(x, x')$ is an arbitrary chain linking $x$ to $x'$, we can minimize \eqref{eqn_stron_triangle_directed} over all such chains maintaining the validity of the inequality,
\begin{equation}\label{eqn_stron_triangle_directed_minimizing}
\tilde{u}_{X}(x, x') \! \leq \! \min_{C(x, x')} \,\, \max_{i|x_i \in C(x, x')} \! A_X(x_i, x_{i+1}) \! = \! \tdu^*_X(x, x'),
\end{equation}
where the last equality is given by the definition of the directed minimum chain cost \eqref{eqn_nonreciprocal_chains}. Thus, the rightmost inequality in \eqref{eqn_inequality_unicity_directed} is proved.

To show the leftmost inequality in \eqref{eqn_inequality_unicity_directed}, consider an arbitrary pair of nodes $x, x' \in X$ and fix $\delta = \tilde{u}^*_X(x, x')$. Then, by Lemma \ref{lemma_axiom_redundancy}, there exists a partition $P_\delta(x,x')=\{B_\delta(x), B_\delta(x')\}$ of the node space $X$ into blocks $B_\delta(x)$ and $B_\delta(x')$ with $x \in B_\delta(x)$ and $x' \in B_\delta(x')$ such that for all points $b \in B_\delta(x)$ and $b' \in B_\delta(x')$ we have
\begin{equation}\label{eqn_partition_proof_quasi_ultrametrics}
A_X(b, b') \geq \delta.
\end{equation}
Focus on a two-node network $N_{u,v}=(\{u,v\}, A_{u,v})$ with $A_{u,v}(u,v)=\delta$ and $A_{u,v}(v,u)=s$ where $s=\sep(X, A_X)$ as defined in \eqref{eqn_def_separation_network}. Denote by $(\{u, v\}, \tilde{u}_{u,v})=\tilde{\ccalH}(N_{u,v})$ the output of applying the method $\tilde{\ccalH}$ to the network $N_{u,v}$. Notice that the map $\phi: X \to \{u,v\}$ such that $\phi(b)=u$ for all $b \in B_\delta(x)$ and $\phi(b')=v$ for all $b' \in B_\delta(x')$ is dissimilarity reducing because, from \eqref{eqn_partition_proof_quasi_ultrametrics}, dissimilarities mapped to dissimilarities equal to $\delta$ in $N_{u,v}$ were originally larger. Moreover, dissimilarities mapped into $s$ cannot have increased due to the definition of separation of a network \eqref{eqn_def_separation_network}. From Axiom (\~A1),
\begin{equation}\label{eqn_two_node_delta_directed}
\tilde{u}_{u,v}(u,v)= A_{u,v}(u,v) = \delta,
\end{equation}
since $N_{u,v}$ is a two-node network. Moreover, since $\phi$ is dissimilarity reducing, from (\~A2) we may assert that
\begin{equation}\label{eqn_transformation_directed_delta}
\tilde{u}_{X}(x,x') \geq \tdu_{u,v}(\phi(x), \phi(x')) = \delta,
\end{equation}
where we used \eqref{eqn_two_node_delta_directed} for the last equality. Recalling that $\tilde{u}^*_X(x, x') = \delta$ and substituting in \eqref{eqn_transformation_directed_delta} concludes the proof of the leftmost inequality in \eqref{eqn_inequality_unicity_directed}.

Since both inequalities in \eqref{eqn_inequality_unicity_directed} hold, we must have $\tdu_X^*(x, x') = \tdu_X(x, x')$ for all $x,x' \in X$. Since this is true for any arbitrary network $N=(X,A_X)$, it follows that the admissible quasi-clustering method must be $\tilde{\ccalH}\equiv\tilde{\ccalH}^*$.
\subsection{The metric on $\mathcal{N}$}\label{sec_metric_on_networks}
Consider two networks $N_X$, $N_Y$ $\in$ $\mathcal{N}$ such that $N_X=(X, A_X)$ and $N_Y=(Y, A_Y)$. A \emph{correspondence} between the sets $X$ and $Y$ is any subset
$R\subseteq X\times Y$ such that $\pi_1(R)=X$ and $\pi_2(R)=Y$. Here,
$\pi_1$ and $\pi_2$ are the usual coordinate-wise projections. The \emph{distortion} $\mathrm{dis}(R)$ of a correspondence $R$ between networks $N_X$ and $N_Y$ is defined  as 
$$\mathrm{dis}(R):=\max_{(x,y),(x',y')\in R}|A_X(x,x')-A_Y(y,y')|.$$
The underlying notion of equality on $\mathcal{N}$ that we use is the following: we say that networks $N_X$ and  $N_Y$
are \emph{isomorphic} or indistinguishable if and only if there exists a
bijection $\phi:X\rightarrow Y$ such that
$A_X(x,x')=A_Y(\phi(x),\phi(x'))$ for all $x,x'\in X$. Given $N_X$ and $N_Y$, we define the \emph{network distance} $d_{\mathcal{N}}$ on $\mathcal{N}\times \mathcal{N}$ as
\begin{equation}\label{eq:dN}
d_{\mathcal{N}}\big(N_X,N_Y\big):=\frac{1}{2}\min_{R}\mathrm{dis}(R),
\end{equation}
where $R$ spans all correspondences between $X$ and $Y$. The structure
of this distance
is similar to that of the Gromov-Hausdorff distance \cite{book-gromov} that is often used in the
context of compact metric spaces. In our context, it
still provides a legitimate distance on the collection $\mathcal{N}$
modulo our chosen notion of isomorphism.

\begin{theorem}\label{thm:dN-metric}
The network distance defined in (\ref{eq:dN}) is a legitimate metric on
$\mathcal{N}$ modulo isomorphism of networks.
\end{theorem}
\begin{myproof}
That $d_{\mathcal{N}}$ is symmetric and non-negative is clear. Assume now that $X$ and
$Y$ are isomorphic and let $\phi:X\rightarrow Y$ be a bijection
providing this isomorphism. Then, consider $R_\phi=\{(x,\phi(x)),\,x\in
X\}$. Since $\phi$ is a bijection, it is easy to check that $R_\phi$ is a
correspondence between $X$ and $Y$. Finally, by definition of $\phi$,
$A_X(x,x')=A_Y(y,y')$ for all $(x,y),(x',y')\in R_\phi$. Hence
$$0\leq\dN{X}{Y}\leq \frac{1}{2}\mathrm{dis}(R_\phi)=0$$
and $\dN{X}{Y}=0$ follows.

The triangle inequality follows from the following observation: if $R$
is a correspondence between $X$ and $Z$ and $S$ is a correspondence
between $Z$ and $Y$, then
\begin{equation}\label{eq:dT}
T := \{(x,y),|\exists z\in Z \text{ with } (x,z)\in R,\:
  (z,y)\in S\}
\end{equation}
is a correspondence between $X$ and $Y$. To show that $T$ is in fact a correspondence, we have to prove that for every $x \in X$ there exists $y \in Y$ such that $(x,y) \in T$. Similarly, we must require that for every $y \in Y$ there exists $x \in X$ such that $(x,y) \in T$. To see this, pick an arbitrary $x \in X$, by definition of $R$, there must exist $z \in Z$ such that $(x,z) \in R$. By definition of $S$, there must exist $y \in Y$ such that $(z, y) \in S$. Hence, there exists $(x, y) \in T$ for every $x \in X$. Similarly, the result can be proven for every element of the set $Y$.

We can prove the triangle inequality in the following way. Consider $R$ and $S$ to be the minimizing correspondences associated with distances $\dN{X}{Z}$ and $\dN{Z}{Y}$ respectively and define $T$ as given by (\ref{eq:dT}). Note that $T$ need not be the minimizing correspondence for $\dN{X}{Y}$. Hence,
\begin{equation}\label{eq:Triangle}
\dN{X}{Y}\leq \frac{1}{2} \mathrm{dis}(T)
\end{equation}
Furthermore, if we add and subtract  $A_Z(z,z')$ within the absolute value defining the distortion of $T$ in (\ref{eq:Triangle}), where $z$ and $z'$ are the elements in the definition of $T$ (\ref{eq:dT}), and we use the fact that the maximum of the absolute value of a sum is less than or equal to the sum of the maximums of absolute values, we obtain
\begin{eqnarray}
&& \dN{X}{Y}\nonumber\\ &\leq& \frac{1}{2} \max_{(x,z),(x',z')\in R}|A_X(x,x')-A_Z(z,z')| \nonumber\\
&+& \frac{1}{2} \max_{(z,y),(z',y')\in S}|A_Z(z,z')-A_Y(y,y')| \label{eq:Triangle2}
\end{eqnarray}
By noting that the expression on the right hand side of (\ref{eq:Triangle2}) is the sum of $\dN{X}{Z}$ and $\dN{Z}{Y}$, the proof of the triangle inequality is completed. 

Finally, the most delicate part of the proof is checking that
$\dN{X}{Y}=0$ implies that $X$ and $Y$ are isomorphic. Assume that $R$
is a correspondence such that $A_X(x,x')=A_Y(y,y')$ for all $(x,y)$
and $(x',y')$ both in $R$. Define $\phi:X\rightarrow Y$ in the
following way: for each $x\in X$ let $Rx\subseteq Y$ be the set of all
$y$ such that $(x,y)\in R$. The fact that $R$ is a correspondence
forces that $Rx\neq \emptyset$. Hence, we can choose \emph{any} $y$ in
$Rx$ and declare $\phi(x)=y$.

Define in the same way a function $\psi:Y\rightarrow X$. Notice that
then we forcibly have that $A_X(x,x')=A_Y(\phi(x),\phi(x'))$ for
all $x,x'\in X$ and also $A_X(\psi(y),\psi(y'))=A_Y(y,y')$ for all
$y,y'\in Y$. 

To prove that $\phi$ is \emph{injective}, assume that
$x\neq x'$ but $\phi(x)=\phi(x'),$ then
$A_X(x,x')=A_Y(\phi(x),\phi(x'))=0$, which contradicts our definition
of networks. In a similar manner one checks that $\psi$ must also be
injective. 

So we have constructed two injections, one from $X$ into $Y$, and one
in the opposite direction. The Cantor-Bernstein-Schroeder theorem now
applies and guarantees that there exists a bijection between $X$ and $Y$. This immediately forces $X$
and $Y$ to have the same cardinality, and in particular, it forces
$\phi$ (and $\psi$) to be bijections. This concludes the proof. 
\end{myproof}

\subsection{Proof of Theorem \ref{thm:stab-dsl}}
Assume $\eta=d_{\mathcal{N}}(N_X,N_Y)$ and let $R$ be a correspondence between $X$ and $Y$ such that $\mathrm{dis}(R)=2\eta$. Write $(X,\tilde{u}_X)= \tilde{\ccalH}^*(N_X)$ and  $(Y,\tilde{u}_Y)= \tilde{\ccalH}^*(N_Y)$. We will prove that $|\tilde{u}_X(x,x')-\tilde{u}_Y(y,y')|\leq 2\eta$ for all $(x,y),(x',y')\in R$ which will imply the claim. Fix $(x,y)$ and $(x',y')$ in $R$. Pick any $x=x_0,x_1,\ldots,x_n=x'$ in $X$ such that $\max_{i}A_X(x_i,x_{i+1})=\tilde{u}_X(x,x').$ Choose $y_0,y_1,\ldots,y_n\in Y$ so that $(x_i,y_i)\in R$ for all $i=0,1,\ldots,n.$ Then, by definition of $\tilde{u}_Y(y,y')$ and the definition of $\eta$:
\begin{eqnarray*}\tilde{u}_Y(y,y')&\leq& \max_iA_Y(y_i,y_{i+1})\\
&\leq& \max_i A_X(x_i,x_{i+1})+2\eta\\
 &=&\tilde{u}_X(x,x')+ 2\eta.
\end{eqnarray*}

By symmetry, one also obtains $\tilde{u}_X(x,x')\leq \tilde{u}_Y(y,y')+2\eta$, and the conclusion follows form the arbitrariness of $(x,y),(x',y')\in R$ and the definition of $d_\mathcal{N}.$
\subsection{Proof of Proposition \ref{prop:scale}}
Fix any $(X,A_X)\in \ccalN$ and write $\tilde{\ccalH}^*(X,A_X) = (X,\tilde{u}_X)$. Pick any change of scale function $\Psi$ and write $(X,\tilde{u}_X^\Psi) = \tilde{\ccalH}^*(X,\Psi(A_X))$. We need to prove that $\tilde{u}^\Psi_X = \Psi(\tilde{u}_X)$. But this follows directly from the explicit structure given in equation (\ref{eqn_nonreciprocal_chains}) and the fact that $\Psi$ is non-decreasing.

\subsection{Further invariances: vertex permutations and the metric closure} \label{sec_inv_supp}

Note that Theorem \ref{thm:stab-dsl} implies that DSL behaves well under permutations of the vertices. The distance between a given network and a second one obtained by permuting its nodes is null. Thus, by Theorem \ref{thm:stab-dsl}, the distance between the corresponding output quasi-dendrograms must be null as well. More precisely, if $(X,A_X)\in\mathcal{N}$, $\tilde{\ccalH}^*(X,A_X)=(X,\tilde{u}_X)$, and  $\varphi:X\rightarrow X$ is any bijection, then $\tilde{\ccalH}^*(X,A_X\circ(\varphi,\varphi)) = (X,\tilde{u}_X\circ(\varphi,\varphi))$.  This means that permuting the labels of points before applying DSL yields the same result as permuting the labels a posteriori.

For any  $(X,A_X)\in\mathcal{N}$ let $\bar{A}_X$ be the \emph{maximal} function satisfying $\bar{A}_X\leq {A}_X$ pointwisely which in addition satisfies the \emph{directed triangle inequality}: $\bar{A}_X(x,x')\leq \bar{A}_X(x,x'')+\bar{A}_X(x'',x')$ for all $x,x',x''\in X$. Then, one can also prove (similar to the proof of Theorem 18 in \cite{clust-um}) that  $\tilde{\ccalH}^*(X,A_X)= \tilde{\ccalH}^*(X,\bar{A}_X)$ for all $X\in\mathcal{N}$.

\subsection{Proof of Proposition \ref{prop_algorithms}}
In Ch.6, Section 6.1 of \cite{GondranMinoux08} it is shown that if $A_X$ is a dissimilarity matrix then its quasi inverse $A_X^*$ in the dioid $(\reals^+\cup\{+\infty\},\min,\max)$ contains information about the minimum infinity norm of chains in the network. In fact, $[A_X^*]_{i,j}$ contains the minimum infinity norm of all the chains connecting node $i$ with node $j$. In \cite{GondranMinoux08}, the analysis is done for the symmetric case but its extension to the asymmetric case is immediate as we present here,
\begin{equation}\label{eqn_quasi_inverse_nonrecip}
[A_X^*]_{i,j} = \min_{C(x_i,x_j)} \,\,\, \max_{k | x_k \in C(x_i,x_j)} \,\, A_X(x_k,x_{k+1}).
\end{equation}
By comparing \eqref{eqn_quasi_inverse_nonrecip} with \eqref{eqn_nonreciprocal_chains}, we can state that 
\begin{equation}\label{eqn_quasi_inverse_tdu} 
A_X^*=\tdu^*_X.
\end{equation}
Hence, if we show that $A^*_X=A^{(n-1)}_X$, then \eqref{eqn_quasi_inverse_tdu} implies \eqref{eqn_algo_quasi_ultrametric}, completing the proof. Recall the quasi inverse $A^*_X$ definition in the dioid $(\reals^+\cup\{+\infty\},\min,\max)$ from Ch. 4, Definition 3.1.2 in \cite{GondranMinoux08}
\begin{equation}\label{eqn_def_quasi_inverse}
A_X^* = \lim_{k\to \infty } I \oplus A_X \oplus A_X^{(2)} \oplus ... \oplus A_X^{(k)},
\end{equation}
where $I$ has zeros in the diagonal and $+ \infty$ in the off diagonal elements.

However, in our dioid algebra where the $\oplus$ operation is idempotent, i.e. $a \oplus a = a$ for all $a$, it can be shown as in Ch. 4, Proposition 3.1.1 in \cite{GondranMinoux08} that 
\begin{equation}\label{eqn_quasi_inverse_develop_0}
I \oplus A_X \oplus A_X^{(2)} \oplus ... \oplus A_X^{(k)}= (I \oplus A_X)^{(k)}.
\end{equation}
In our case, it is immediate that $I \oplus A_X = A_X$, since diagonal elements are null in both matrices and the off diagonal elements in $I$ are $+\infty$. Hence, the minimization operation $\oplus$ preserves $A_X$. Consequently, \eqref{eqn_quasi_inverse_develop_0} becomes
\begin{equation}\label{eqn_quasi_inverse_develop}
I \oplus A_X \oplus A_X^{(2)} \oplus ... \oplus A_X^{(k)}= A_X^{(k)}.
\end{equation}
Taking the limit to infinity in both sides of equality \eqref{eqn_quasi_inverse_develop} and using the quasi inverse definition \eqref{eqn_def_quasi_inverse}, we get
\begin{equation}\label{eqn_quasi_inverse_develop_2}
A_X^*= \lim_{k \to \infty} A_X^{(k)}.
\end{equation}
Finally, it can be shown as in Theorem 1 of Ch.4, Section 3.3 in \cite{GondranMinoux08} that $A_X^{(n-1)}=A_X^{(n)}$, proving that the limit in \eqref{eqn_quasi_inverse_develop_2} is well defined and, more importantly, that $A_X^*=A_X^{(n-1)}$, as wanted.

\subsection{Applications}\label{sec_sup_mat_applications}

The dissimilarity function $A_S$ of the migration network $N_S$ used in Section \ref{sec_applications} of the paper is computed as follows. Denote by $M: S \times S \to \reals_+$ the migration flow function given by the U.S. census bureau in which $M(s, s')$ is the number of individuals that migrated from state $s$ to $s'$ in year 2011 and $M(s, s)=0$ for all $s, s' \in S$. We then construct the asymmetric network $N_S=(S, A_S)$ with node set $S$ and dissimilarities $A_S$ such that $A_S(s, s)=0$ for all $s \in S$ and
\begin{equation}\label{eqn_def_migration_dissimilarity}
   A_S(s, s') = f \left( \frac{M(s, s')}{\sum_i M(s_i, s')}\right),
\end{equation}
for all $s \neq s' \in S$ where $f: [0, 1) \to \reals_{++}$ is a given decreasing function. The normalization $M(s, s')/\sum_i M(s_i, s')$ in \eqref{eqn_def_migration_dissimilarity} can be interpreted as the probability that an immigrant to state $s'$ comes from state $s$. The role of the decreasing function $f$ is to transform the similarities $M(s, s')/\sum_i M(s_i, s')$ into corresponding dissimilarities. For the experiments here we use $f(x) = 1-x$. However, due to the scale invariance property of DSL [cf. Proposition \ref{prop:scale}] , the particular form of $f$ is of little consequence to our analysis. Indeed, the influence structure between blocks of states obtained when quasi-clustering the network $N_S$ is independent of the particular choice of the decreasing function $f$.

\begin{figure*}
\hspace{-0.6in}
\includegraphics[width=1.15 \linewidth, height=0.3 \linewidth]
                {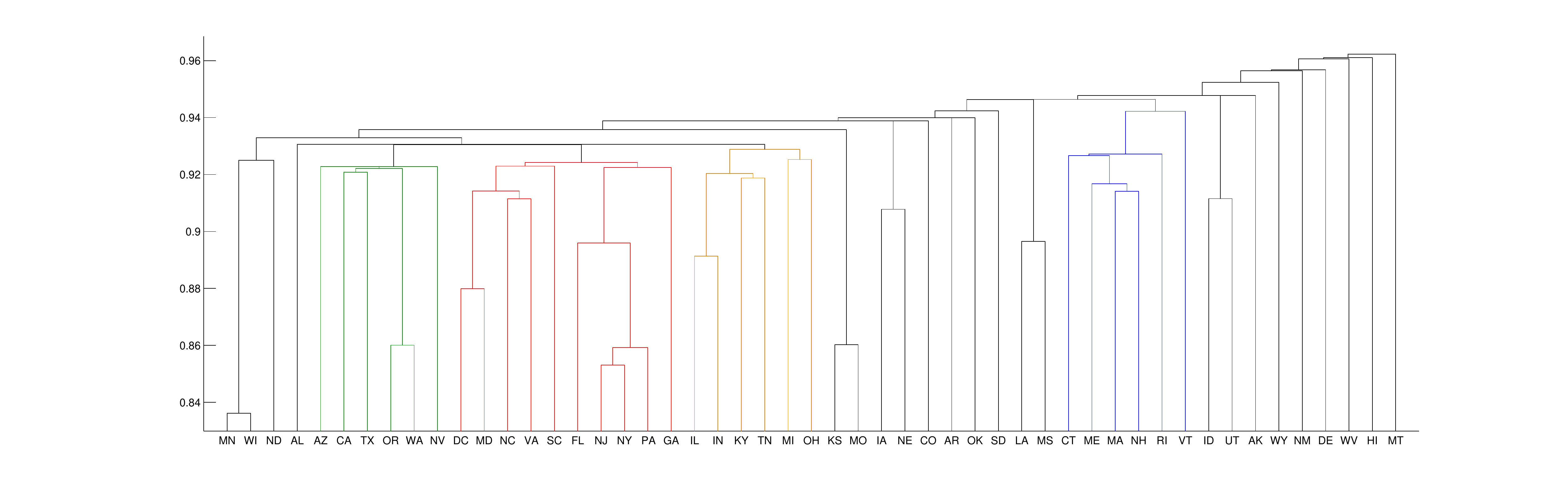}
\caption{Dendrogram component $D^*_S$ of the quasi-dendrogram $\tilde{D}^*_S= (D^*_S, E^*_S)$. The clustering of states is highly influenced by geographical proximity.}
\label{fig_nonreciprocal_us_dendrogram}
\end{figure*}

In Fig. \ref{fig_nonreciprocal_us_dendrogram} we present the dendrogram component $D^*_S$ of the quasi-dendrogram $\tilde{D}^*_S= (D^*_S, E^*_S)$ analyzed in Section \ref{sec_applications}. Some identifiable clusters are highlighted in color to illustrate the influence of geographical proximity in migrational preference. E.g., the blue cluster corresponds to the six states in the region of New England, the red cluster contains the remaining East Coast states with the exception of Delaware, and the green cluster corresponds to states in an extended West Coast plus Texas.

As a second illustrative example of the DSL method, we quasi-cluster a network that records interactions between sectors of the economy. 
The Bureau of Economic Analysis of the U.S. Department of Commerce publishes a yearly table of inputs and outputs organized by economic sectors \cite{USinputoutput}. This table records how economic sectors interact to generate gross domestic product. We focus on the section of uses of this table which shows the inputs to production. More precisely, we are given a set $I$ of 61 industrial sectors as defined by the North American Industry Classification System (NAICS) and a function $U: I \times I \to \reals_+$ where $U(i, i')$ for all $i, i' \in I$ represents how much of the production of sector $i$, expressed in dollars, is used as an input of sector $i'$. The function $U$ should be interpreted as a measure of directed closeness between two sectors. Thus, we define the network of uses $N_I=(I, A_I)$ where the dissimilarity function $A_I$ satisfies $A_I(i,i)=0$ and, for $i \neq i' \in I$, is given by
\begin{equation}\label{eqn_def_io_dissimilarity}
A_I(i, i') = f \left( \frac{U(i, i')}{\sum_k U(i_k, i')}\right),
\end{equation}
where $f: [0, 1) \to \reals_{++}$ is a given decreasing function. The normalization $U(i, i')/\sum_k U(i_k, i')$ in \eqref{eqn_def_io_dissimilarity} can be interpreted as the probability that an input dollar to productive sector $i'$ comes from sector $i$. In this way, we focus on the combination of inputs of a sector rather than the size of the economic sector itself. That is, a small dissimilarity from sector $i$ to sector $i'$ implies that sector $i'$ highly relies on the use of sector $i$ output as an input for its own production. Notice that $U(i,i)$ for $i \in I$ is generally positive, i.e., a sector uses outputs of its own production as inputs in other processes. Consequently, if for a given sector we sum the input proportion from every other sector, we obtain a number less than 1. The role of the decreasing function $f$ is to transform the similarities $U(i, i')/\sum_k U(i_k, i')$ into corresponding dissimilarities. As in the previous application, we use $f(x) = 1-x$, though the particular form of $f$ is of little consequence to the analysis since DSL is scale invariant [cf. Proposition \ref{prop:scale}]. 

%
\begin{figure*}
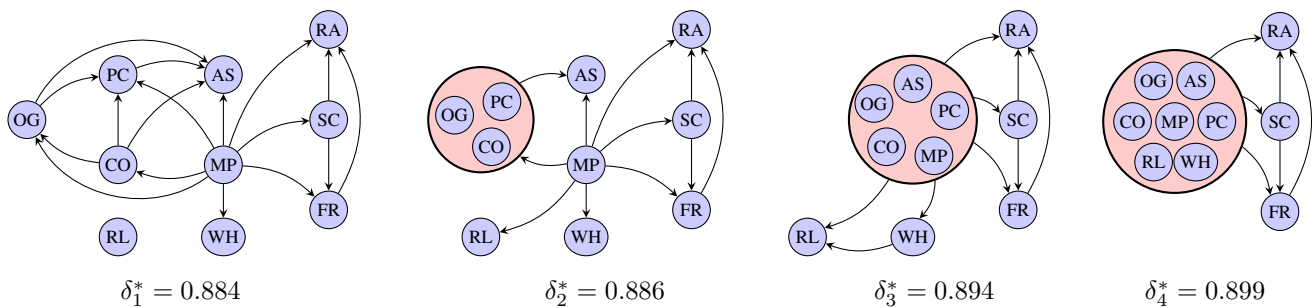

\centering
\centerline{\def \thisplotscale {0.5}
\def \unit {\thisplotscale cm}
\tikzstyle {blue vertex here} = [blue vertex, 
                                 minimum width = 0.7*\unit, 
                                 minimum height = 0.7*\unit, 
                                 anchor=center]
{\begin{tikzpicture}[thick, x = 1.2*\unit, y = 0.96*\unit]

       \node[anchor=south west,inner sep=0] at (-8,0) {\centering\input{quasi-clustering_io_network_1_2.tex}};
        \node[anchor=south west,inner sep=0] at (1.3,0) {\centering\input{quasi-clustering_io_network_2_2.tex}};
         \node[anchor=south west,inner sep=0] at (9.3,0) {\centering\input{quasi-clustering_io_network_3_2.tex}};
          \node[anchor=south west,inner sep=0] at (15.5,0.7) {\centering\input{quasi-clustering_io_network_4_2.tex}};

\node at (-3.7,-1) {$\delta^*_1=0.884$};
\node at (5.7, -1) {$\delta^*_2=0.886$};
\node at (13, -1) {$\delta^*_3=0.894$};
\node at (19, -1) {$\delta^*_4=0.899$};

\end{tikzpicture}} }
\caption{Directed single linkage quasi-clustering method applied to a portion of the sectors of the economy. The edges define a partial order among the blocks of every quasi-partition.}
\label{fig_quasi-dendrogram_example_io_2}
\end{figure*}

The outcome of applying the DSL quasi-clustering method $\tilde{\ccalH}^*$ with output quasi-ultrametrics defined in \eqref{eqn_nonreciprocal_chains} to the network $N_I$ is computed with the algorithmic formula in \eqref{eqn_algo_quasi_ultrametric}. As we did with the migration network, in order to facilitate understanding we present quasi-partitions obtained by restricting the output quasi-ultrametric to a subset of nodes. In Fig. \ref{fig_quasi-dendrogram_example_io_2} we present four quasi-partitions focusing on ten economic sectors; see Table \ref{table_industrial_sectors}. We present quasi-partitions $\tdD^*_I(\delta)$ for four different resolutions $\delta^*_1=0.884$, $\delta^*_2=0.886$, $\delta^*_3=0.894$, and $\delta^*_4=0.899$.

\begin{table}
\small
\centering
\caption{Code and description of industrial sectors}
\begin{tabular}{| c | l| }
  \hline    
 Code & Industrial Sector \\ \hline \hline                    
OG & Oil and gas extraction \\ \hline
CO & Construction \\ \hline
PC & Petroleum and coal products \\ \hline
WH & Wholesale trade \\ \hline
FR & Federal Reserve banks and credit intermediation \\ \hline
SC & Securities, commodity contracts, and investments \\ \hline
RA & Real estate \\ \hline
RL & Rental and leasing serv. and lessors of intang. assets \\ \hline
MP & Misc. professional, scientific, and technical services \\ \hline
AS & Administrative and support services \\ \hline
\end{tabular}
\label{table_industrial_sectors}
\end{table}
The edge component $E^*_I$ of the quasi-dendrogram $\tdD^*_I$ captures the asymmetric influence between clusters. E.g. in the quasi-partition in Fig. \ref{fig_quasi-dendrogram_example_io_2} for resolution $\delta^*_1=0.884$ every cluster is a singleton since the resolution is smaller than that of the first merging. However, the influence structure reveals an asymmetry in the dependence between the economic sectors. At this resolution the professional service sector MP has influence over every other sector except for the rental services RL as depicted by the eight arrows leaving the MP sector. No sector has influence over MP at this resolution since this would imply, except for RL, the formation of a non-singleton cluster. The influence of MP reaches primary sectors as OG, secondary sectors as PC and tertiary sectors as AS or SC. The versatility of MP's influence can be explained by the diversity of services condensed in this economic sector, e.g. civil engineering and architectural services are demanded by CO, production engineering by PC and financial consulting by SC. For the rest of the influence pattern, we can observe an influence of CO over OG mainly due to the construction and maintenance of pipelines, which in turn influences PC due to the provision of crude oil for refining. Thus, from the transitivity (QP2) property of quasi-partitions we have an influence edge from CO to PC. The sectors CO, PC and OG influence the support service sector AS. Moreover, the service sectors RA, SC and FR have a totally hierarchical influence structure where SC has influence over the other two and FR has influence over RA. Since these three nodes remain as singleton clusters for the resolutions studied, the influence structure described is preserved for higher resolutions as it should be from the influence hierarchy property of the edge set $E_S^*(\delta)$ stated in condition (\~D3) in the definition of quasi-dendrogram in Section \ref{sec_quasi_dendrograms}.

At resolution $\delta^*_2=0.886$, we see that the sectors OG-PC-CO have formed a three-node cluster depicted in red that influences AS. At this resolution, the influence edge from MP to RL appears and, thus, MP gains influence over every other cluster in the quasi-partition including the three-node cluster. At resolution $\delta=0.887$ the service sectors AS and MP join the cluster OG-PC-CO and for $\delta^*_3=0.894$ we have this five-node cluster influencing the other five singleton clusters plus the mentioned hierarchical structure among SC, FR, and RA and an influence edge from WH to RL. When we increase the resolution to $\delta^*_4=0.899$ we see that RL and WH have joined the main cluster that influences the other three singleton clusters. If we keep increasing the resolution, we would see at resolution $\delta=0.900$ the sectors SC and FR joining the main cluster which would have influence over RA the only other cluster in the quasi-partition. Finally, at resolution $\delta=0.909$, RA joins the main cluster and the quasi-partition contains only one block.

The influence structure between clusters at any given resolution defines a partial order. More precisely, for every resolution $\delta$, the edge set $E_I^*(\delta)$ defines a partial order between the blocks given by the partition $D^*_I(\delta)$. We can use this partial order to evaluate the relative importance of different clusters by stating that more important sectors have influence over less important ones. E.g., at resolution $\delta^*_1=0.884$ we have that MP is more important than every other sector except for RL, which is incomparable at this resolution. There are three totally ordered chains that have MP as the most important sector at this resolution. The first one contains five sectors which are, in decreasing order of importance, MP, CO, OG, PC, and AS.  The second one is comprised of MP, SC, FR, and RA and the last one only contains MP and WH. At resolution $\delta^*_2=0.886$ we observe that the three-node cluster OG-PC-CO, although it contains more nodes than any other cluster, it is not the most important of the quasi-partition. Instead, the singleton cluster MP has influence over the three-node cluster and, on top of that, is comparable with every other cluster in the quasi-partition. From resolution $\delta^*_3=0.894$ onwards, after MP joins the red cluster, the cluster with the largest number of nodes coincides with the most important of the quasi-partition. At resolution $\delta^*_4=0.899$ we have a total order among the four clusters of the quasi-partition. This is not true for the other three depicted quasi-partitions.   

%
\begin{figure*}
\centering
\centerline{\input{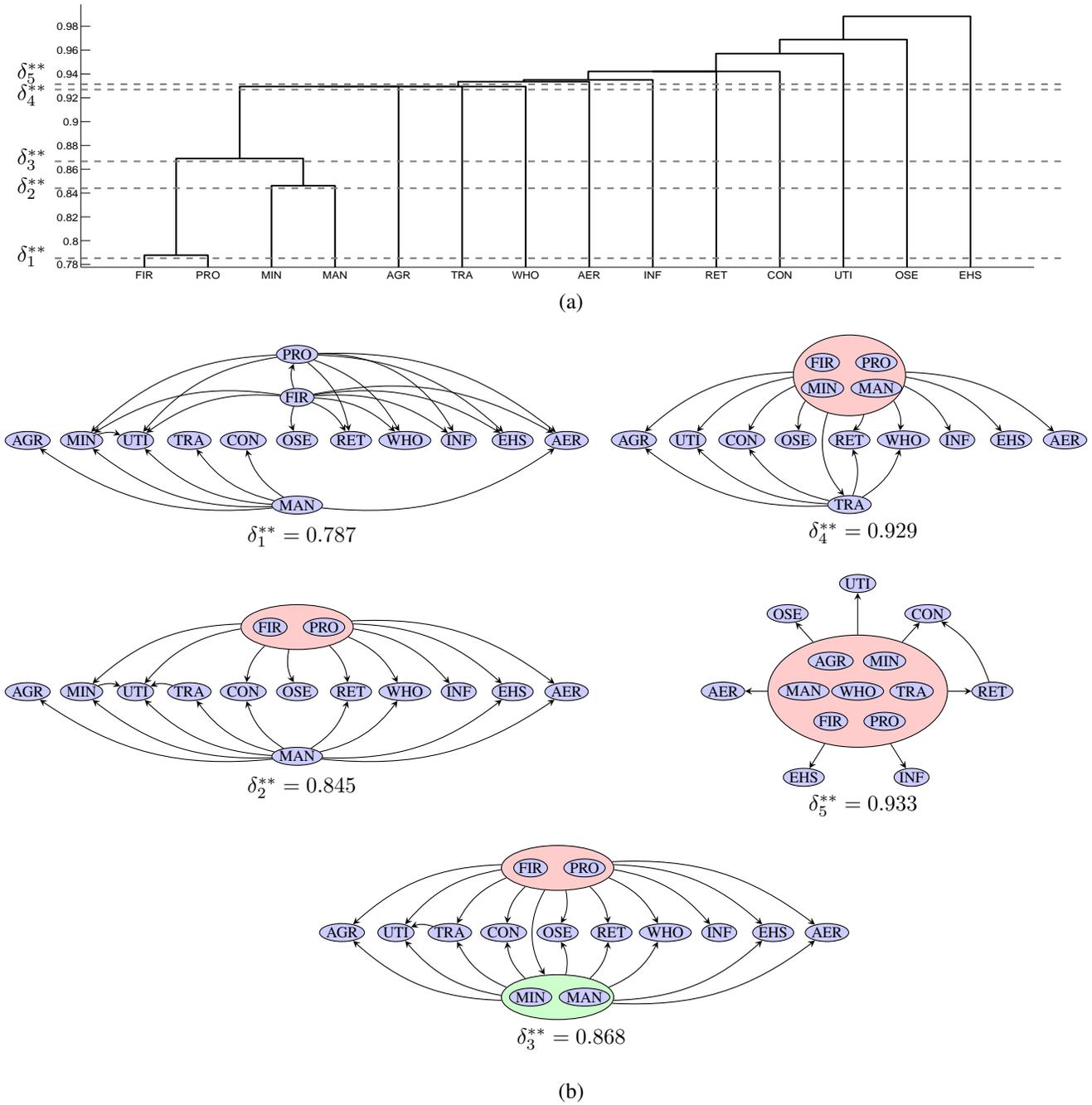} }
\caption{(a) Dendrogram component $D_C^*$ of the quasi-dendrogram $\tdD_C^*=(D_C^*, E_C^*)$. Output of the DSL quasi-clustering method $\tilde{\ccalH}^*$ when applied to the network $N_{C}$. (b) Quasi-partitions. Given by the specification of the quasi-dendrogram $\tdD_C^*$ at a particular resolution $\tdD_C^*(\delta^{**}_k)$ for $k=1, \ldots, 5$.}
\label{fig_quasi-clustering_example_io}
\end{figure*}

As a further illustration of the quasi-clustering method $\tilde{\ccalH}^*$, we apply this method to the network $N_{C}=(C, A_C)$ of consolidated industrial sectors \cite{USinputoutput} where $|C|=14$ -- see Table \ref{table_consolidated_industrial_sectors} -- instead of the original 61 sectors. Dissimilarity function $A_C$ is analogous to $A_I$ but computed for the consolidated sectors. Of the output quasi-dendrogram $\tdD^*_C=(D^*_C, E^*_C)$, in Fig. \ref{fig_quasi-clustering_example_io}-(a) we show the dendrogram component $D^*_C$ and in Fig. \ref{fig_quasi-clustering_example_io}-(b) we depict the quasi-partitions $\tdD^*_C(\delta^{**}_i)$ for $\delta^{**}_1=0.787$, $\delta^{**}_2=0.845$, $\delta^{**}_3=0.868$, $\delta^{**}_4=0.929$, and $\delta^{**}_5=0.933$. The reason we use the consolidated network $N_C$ is to facilitate the visualization of quasi-partitions that capture every sector of the economy instead of only ten particular sectors as in the previous application.

\begin{table}
\small
\centering
\caption{Code and description of consolidated industrial sectors}
\begin{tabular}{| c | l| }
  \hline    
 Code & Consolidated Industrial Sector \\ \hline \hline                    
AGR & Agriculture, forestry, fishing, and hunting \\ \hline
MIN & Mining \\ \hline
UTI & Utilities \\ \hline
CON & Construction \\ \hline
MAN & Manufacturing \\ \hline
WHO & Wholesale trade \\ \hline
RET & Retail trade \\ \hline
TRA & Transportation and warehousing \\ \hline
INF & Information \\ \hline
FIR & Finance, insurance, real estate, rental, and leasing \\ \hline
PRO & Professional and business services \\ \hline
EHS & Educational services, health care, and social assistance \\ \hline
AER & Arts, entertain., recreation, accomm., and food serv. \\ \hline
OSE & Other services, except government \\ \hline
\end{tabular}
\label{table_consolidated_industrial_sectors}
\end{table}

The quasi-dendrogram $\tdD^*_C$ captures the asymmetric influences between clusters of industrial sectors at every resolution. E.g., at resolution $\delta^{**}_1=0.787$ the dendrogram $D^*_C$ in Fig. \ref{fig_quasi-clustering_example_io}-(a) informs us that every industrial sector forms its own singleton cluster. However, this simplistic representation, characteristic of clustering methods, ignores the asymmetric relations between clusters at resolution $\delta^{**}_1$. These influence relations are formalized in the quasi-dendrogram $\tdD^*_C$ with the introduction of the edge set $E^*_C(\delta)$ for every resolution $\delta$. In particular, for $\delta^{**}_1$ we see in Fig. \ref{fig_quasi-clustering_example_io}-(b) that the sectors of `Finance, insurance, real estate, rental, and leasing' (FIR) and `Manufacturing' (MAN) combined have influence over the remaining 12 sectors. More precisely, the influence of FIR is concentrated on the service and commercialization sectors of the economy whereas the influence of MAN is concentrated on primary sectors, transportation, and construction. Furthermore, note that due to the transitivity (QP2) property of quasi-partitions defined in Section \ref{sec_full_characterization_asymmetric}, the influence of FIR over `Professional and business services' (PRO) implies influence of FIR over every sector influenced by PRO. The influence among the remaining 11 sectors, i.e. excluding MAN, FIR and PRO, is minimal, with the `Mining' (MIN) sector influencing the `Utilities' (UTI) sector. This influence is promoted by the influence of the `Oil and gas extraction' (OG) subsector of MIN over the utilities sector. At resolution $\delta^{**}_2=0.845$, FIR and PRO form one cluster, depicted in red, and they add an influence to the `Construction' (CON) sector apart from the previously formed influences that must persist due to the influence hierarchy property of the edge set $E_C^*(\delta)$ stated in condition (\~D3) in the definition of quasi-dendrogram in Section \ref{sec_quasi_dendrograms}. The manufacturing sector also intensifies its influences by reaching the commercialization sectors `Retail trade' (RET) and `Wholesale trade' (WHO) and the service sector `Educational services, health care, and social assistance' (EHS). The influence among the rest of the sectors is still scarce with the only addition of the influence of `Transportation and warehousing' (TRA) over UTI. At resolution $\delta^{**}_3=0.868$ we see that mining MIN and manufacturing MAN form their own cluster, depicted in green. The previously formed red cluster has influence over every other cluster in the quasi-partition, including the green one. At resolution $\delta^{**}_4=0.929$, the red and green clusters become one, composed of four original sectors. Also, the influence of the transportation TRA sector over the rest is intensified with the appearance of edges to the primary sector `Agriculture, forestry, fishing, and hunting' (AGR), the construction CON sector and the commercialization sectors RET and WHO. Finally, at resolution $\delta^{**}_5=0.933$ there is one clear main cluster depicted in red and composed of seven sectors spanning the primary, secondary, and tertiary sectors of the economy. This main cluster influences every other singleton cluster. The only other influence in the quasi-partition $\tdD^*_C(0.933)$ is the one of RET over CON. For increasing resolutions, the singleton clusters join the main red cluster until at resolution $\delta=0.988$ the 14 sectors form one single cluster.

The influence structure at every resolution induces a partial order in the blocks of the corresponding quasi-partition. As done in previous examples, we can interpret this partial order as a relative importance ordering. E.g., we can say that at resolution $\delta^{**}_1=0.787$, MAN is more important that MIN which in turn is more important than UTI which is less important than PRO. However, PRO and MAN are not comparable at this resolution. At resolution $\delta^{**}_4=0.929$, after the red and green clusters have merged together at resolution $\delta=0.869$, we depict the combined cluster as red. This representation is not arbitrary, the red color of the combined cluster is inherited from the most important of the two component cluster. The fact that the red cluster is more important than the green one is represented by the edge from the former to the latter in the quasi-partition at resolution $\delta^{**}_3$. In this sense, the edge component $E^*_C$ of the quasi-dendrogram formalizes a hierarchical structure between clusters at a fixed resolution apart from the hierarchical structure across resolutions given by the dendrogram component $D^*_C$ of the quasi-dendrogram. E.g., if we focus only on the dendrogram $D^*_C$ in Fig. \ref{fig_quasi-clustering_example_io}-(a), the nodes MIN and MAN seem to play the same role. However, when looking at the quasi-partitions at resolutions $\delta^{**}_1$ and $\delta^{**}_2$, it is clear that MAN has influence over a larger set of nodes than MIN and hence plays a more important role in the clustering for increasing resolutions. Indeed, if we delete the three nodes with the strongest influence structure, namely PRO, FIR, and MAN, and apply the quasi-clustering method $\tilde{\ccalH}^*$ on the remaining 11 nodes, the first merging occurs between the mining MIN and utilities UTI sectors at $\delta=0.960$. At this same resolution, in the original dendrogram component in Fig. \ref{fig_quasi-clustering_example_io}-(a), a main cluster composed of 12 nodes only excluding `Other services, except government' (OSE) and EHS is formed. This indicates that by removing influential sectors of the economy, the tendency to cluster of the remaining sectors is decreased.

\end{appendices}


\end{document}